\documentclass[10pt,twocolumn,letterpaper]{article}

\usepackage{cvpr}              

\usepackage[dvipsnames]{xcolor}

\definecolor{cvprblue}{rgb}{0.21,0.49,0.74}
\usepackage[pagebackref,breaklinks,colorlinks,citecolor=cvprblue]{hyperref}
\usepackage{algorithm}
\usepackage{algpseudocode}
\usepackage{multirow}
\usepackage{makecell}
\usepackage{tikz}

\def\paperID{15103} 
\def\confName{CVPR}
\def\confYear{2024}

\title{FastBlend: a Powerful Model-Free Toolkit Making Video Stylization Easier}

\author{Zhongjie Duan$^1$, Chengyu Wang$^2$, Cen Chen$^1$, Weining Qian$^1$, Jun Huang$^2$, Mingyi Jin$^3$\\
$^1$East China Normal University, Shanghai, China\\
$^2$Alibaba Group, Hangzhou, China\\
$^3$Individual Researcher, Shanghai, China\\
{\tt\small zjduan@stu.ecnu.edu.cn, chengyu.wcy@alibaba-inc.com, cenchen@dase.ecnu.edu.cn,}\\
{\tt\small huangjun.hj@alibaba-inc.com, wnqian@dase.ecnu.edu.cn, jinmingyi1998@sina.cn}
}

\begin{document}
\maketitle
\begin{abstract}
With the emergence of diffusion models and rapid development in image processing, it has become effortless to generate fancy images in tasks such as style transfer and image editing. However, these impressive image processing approaches face consistency issues in video processing. In this paper, we propose a powerful model-free toolkit called FastBlend to address the consistency problem for video processing. Based on a patch matching algorithm, we design two inference modes, including blending and interpolation. In the blending mode, FastBlend eliminates video flicker by blending the frames within a sliding window. Moreover, we optimize both computational efficiency and video quality according to different application scenarios. In the interpolation mode, given one or more keyframes rendered by diffusion models, FastBlend can render the whole video. Since FastBlend does not modify the generation process of diffusion models, it exhibits excellent compatibility. Extensive experiments have demonstrated the effectiveness of FastBlend. In the blending mode, FastBlend outperforms existing methods for video deflickering and video synthesis. In the interpolation mode, FastBlend surpasses video interpolation and model-based video processing approaches. The source codes have been released on GitHub\footnote{https://github.com/Artiprocher/sd-webui-fastblend}.
\end{abstract}

\section{Introduction}

In recent years, there has been a rapid development in the field of image processing. Notably, the diffusion models \cite{saharia2022photorealistic, ramesh2022hierarchical} trained on massive datasets have ushered in a transformative era in image synthesis. It has been demonstrated that diffusion models outperform Generative Adversarial Networks (GANs) \cite{goodfellow2014generative} comprehensively \cite{dhariwal2021diffusion}, and even reach the level of creative ability comparable to that of human artists \cite{yang2022diffusion}. Stable Diffusion \cite{rombach2022high} has become the most popular model architecture in open-source communities and has been applied to various domains. In tasks such as image style transfer \cite{mou2023t2i}, super-resolution \cite{li2022srdiff}, and image editing \cite{hertz2022prompt}, diffusion-based approaches have achieved noteworthy milestones in success.

However, when we extend these image processing techniques to video processing, we face the issue of maintaining video consistency \cite{yu2021generating, lei2023blind, yang2023rerender}. Since each frame in a video is processed independently, the direct application of image processing methods typically results in incoherent contents, leading to noticeable flickering in generated videos. In recent years, numerous approaches have been proposed to enhance the consistency of generated videos. We summarize these video processing methods as follows: \textbf{1) Full Video Rendering} \cite{wu2023tune, ceylan2023pix2video}: Each frame is processed through diffusion models. To enhance frame consistency, specialized mechanisms designed for video processing are employed. \textbf{2) Keyframe Sequence Rendering} \cite{singer2022make, yang2023rerender}: A sequence of keyframes is processed by diffusion models, while interpolation methods are leveraged to generate the remaining frames. \textbf{3) Single-Frame Rendering} \cite{ouyang2023codef}: Only a single frame is processed through diffusion models, and the complete video is subsequently rendered according to the motion information extracted from the original video. Currently, these approaches still face challenges, as highlighted in a recent survey \cite{xing2023survey}. In full video rendering, existing methods struggle to ensure video coherence, and the noticeable flickering still exists in some cases. In keyframe sequence rendering, the content in adjacent keyframes remains inconsistent, resulting in abrupt and unnatural transitions in non-keyframes. In single-frame rendering, due to the limited information in a single frame, frame tearing is commonly observed in high-speed motion videos.

Undoubtedly, in recent years, some zero-shot methods \cite{qi2023fatezero, ceylan2023pix2video, khachatryan2023text2video} have improved video consistency by modifying the generation process of diffusion models. However, there is still potential for further enhancement in the performance of these methods. For example, we can combine diffusion models with other advanced video processing techniques \cite{ouyang2023codef, barnes2009patchmatch, lei2023blind} to construct a powerful pipeline. In this paper, we propose a powerful model-free toolkit called FastBlend. To maximize the compatibility with existing methods, we exclusively operate in the image space rather than the latent space \cite{duan2023diffsynth}, avoiding modifications to diffusion models. In other words, FastBlend can work as a post-processor in a video-to-video translation pipeline. This toolkit supports all three aforementioned kinds of synthesis approaches. It can transform an incoherent video into a fluent and realistic one by blending multiple frames, and it can also render the entire video based on one or more keyframes. In the human evaluation, participants unanimously found that FastBlend’s overall performance is significantly better than the baseline methods. Furthermore, we implement FastBlend with a focus on highly parallel processing on GPUs \cite{luebke2008cuda}, achieving exceptional computational efficiency. Running on an NVIDIA 3060 laptop GPU, FastBlend can transform 200 flickering frames into a fluent video within only 8 minutes. The contributions of this paper can be summarized as follows:
\begin{itemize}
  \item We propose FastBlend, a powerful toolkit capable of generating consistent videos, while remaining compatible with most diffusion-based image processing methods.
  \item We boldly employ a model-free patch matching algorithm to effectively align content within videos, thereby enabling precise and rapid object tracking for frame blending and interpolation.
  \item We design an efficient algorithmic architecture, including several compiling-optimized kernel functions and tree-like data structures for frame blending. This architectural design leads to remarkable computational efficiency.
\end{itemize}

\section{Related Work}

\subsection{Image Processing}

Image processing encompasses downstream tasks such as image style transfer, super-resolution, and image editing. Stable Diffusion \cite{rombach2022high}, trained on a large-scale text-image dataset \cite{schuhmann2022laion}, has emerged as a powerful backbone in academic communities. Image processing methods based on Stable Diffusion have achieved impressive success. For instance, by leveraging ControlNet \cite{zhang2023adding} and T2I-Adapter \cite{mou2023t2i}, it is possible to redraw appearances while preserving the underlying image structure, or render hand-drawn sketches into realistic photographs. Textual Inversion \cite{gal2022image}, LoRA \cite{hu2021lora}, and DreamBooth \cite{ruiz2023dreambooth} provide the flexibility to fine-tune Stable Diffusion for generating some specific objects. In the domain of image editing, approaches such as Prompt-to-Prompt \cite{hertz2022prompt}, SDEdit \cite{meng2021sdedit}, and InstructPix2Pix \cite{brooks2023instructpix2pix} are capable of editing images according to user input in the form of text or sketches. Besides diffusion-based approaches, Real-ESRGAN \cite{wang2021real}, CodeFormer \cite{zhou2022towards} and other image super-resolution and restoration methods can be combined with diffusion models to create many fancy image processing pipelines. These image processing methods have inspired subsequent advancements in video processing.

\subsection{Video Processing}

Different from image processing, video processing poses more challenges, typically requiring more computational resources while ensuring the consistency of videos. The recent trend in research focuses on extending image diffusion models to video processing. For instance, Gen-1 \cite{esser2023structure} introduces temporal structures into a diffusion model and trains it to restyle videos. Training a video diffusion model is a costly endeavor, leading some researchers to explore zero-shot video processing methods based on the diffusion models in open-source communities. Notable examples of zero-shot video processing methods include Text2LIVE \cite{bar2022text2live}, FateZero \cite{qi2023fatezero}, Pix2Video \cite{ceylan2023pix2video}, and Text2Video-Zero \cite{khachatryan2023text2video}, among others. These zero-shot video processing methods process videos frame by frame, which demands substantial computational resources and poses challenges in maintaining video consistency. To address these issues, approaches like Make-A-Video \cite{singer2022make} and Rerender-A-Video \cite{yang2023rerender} employ keyframe rendering and video interpolation \cite{jamrivska2019stylizing} to enhance video consistency, while CoDeF \cite{ouyang2023codef} aims to render an entire video using only a single keyframe. The trade-off on the number of keyframes rendered with diffusion models becomes a significant consideration in current video processing tasks. More keyframes result in higher video quality, but lead to lower consistency.

\section{Methodology}

\subsection{Overview}

FastBlend is a model-free toolkit that supports easy deployment, as it does not require training. Based on a patch matching algorithm expounded in Section \ref{subsection:patch_matching} and an image remapping algorithm expounded in Section \ref{subsection:remapping}, FastBlend provides two functions, namely blending and interpolation, which will be discussed in Section \ref{subsection:blending} and Section \ref{subsection:interpolate}.

\subsection{Patch Matching}
\label{subsection:patch_matching}

\begin{algorithm}
  \small
  \caption{Base patch matching algorithm}
  \label{algorithm:base_patch_matching}
  \begin{algorithmic}
    \Statex \textbf{Input:} $S\in \mathbb{R}^{h\times w\times 3}$: source image
    \Statex \textbf{Input:} $T\in \mathbb{R}^{h\times w\times 3}$: target image 
    \Statex \textbf{Input:} $\mathcal L$: customizable loss function 
    \Statex \textbf{Input:} $n$: number of iterations
    \State Randomly initialize $F\in \mathbb{N}^{h\times w\times 2}$
    \For{each pyramid level $(h', w')$}
      \State Resize images $S,T$ to $S',T'\in \mathbb{R}^{h'\times w'\times 3}$
      \State Upsample $F$ to $\mathbb{R}^{h'\times w'\times 2}$
      \State Initialize error matrix $E \gets \mathcal L(S', T', F)\in \mathbb{R}^{h\times w}$.
      \For{$i=1,2,\cdots,n$}
        \For{$F'$ in updating sequence of $F$}
          \State $E' \gets \mathcal L(S', T', F')$
          \State $F(E'<E) \gets F'(E'<E)$
          \State $E(E'<E) \gets E'(E'<E)$
        \EndFor
      \EndFor
    \EndFor
    \State \Return $F$
  \end{algorithmic}
\end{algorithm}

Given a source image $S\in \mathbb{R}^{h\times w\times 3}$ and a target image $T\in \mathbb{R}^{h\times w\times 3}$, we compute an approximate NNF (nearest-neighbor field \cite{mount2010ann}) $F=\text{NNF}(S,T) \in \mathbb{N}^{h\times w\times 2}$ which represents the matches between the two images. NNF is proposed for image reconstruction \cite{guo2021image, liu2021editing}, not for motion estimation, thus it is different from the optical flow \cite{teed2020raft}. For convenience, we use $S[x, y]\in \mathbb R^{(2p+1)\times(2p+1)}$ to denote the patch which centers around the position $(x, y)$, and use $S(x, y)\in \mathbb R^3$ to denote the pixel at $(x, y)$. More precisely, $F(i, j)=(x, y)$ denotes that the patch $T[i, j]$ matches $S[x, y]$. The pseudocode of the base patch matching algorithm is presented in Algorithm \ref{algorithm:base_patch_matching}. In this algorithm, an image pyramid is constructed. The NNF is first estimated with low resolution and then upscaled for fine-tuning. We use a customizable loss function $\mathcal L$ to calculate the errors of matches. The base loss function is formulated as
\begin{equation}
  \mathcal L(S,T,F)_{x,y}=||S[F(x,y)]-T[x,y]||_2^2.
\end{equation}
We also design several customized loss functions, which are described detailedly in \ref{subsection:blending} and \ref{subsection:interpolate}. The estimated NNF $F$ is updated iteratively. In each iteration, we scan the updating sequence of $F$ and replace the values which can reduce the error. The updating sequence is generated by the following two steps, which are proposed by Barnes et al \cite{barnes2009patchmatch}.

\begin{itemize}
  \item \textbf{Propagation}: update matches using adjacent matches. $F'(x,y)=F(x+d_x,y+d_y)-(d_x,d_y)$, where $(d_x,d_y)\in \{(-1, 0), (1, 0), (0, -1), (0, 1)\}$ corresponds to the four directions in the image. 
  \item \textbf{Random search}: search for better matches in the whole image. $F'(x,y)=F(x,y)+(d_x,d_y)$, where $(d_x,d_y)\sim \mathcal U(0, r)$ and $r$ declines exponentially to zero.
\end{itemize}

In the original algorithm \cite{barnes2009patchmatch}, $F$ is updated in a specific manner, i.e., from top to bottom and from left to right. This algorithm can leverage the updated results during an iteration, speeding up the convergence. However, this algorithm is not parallelizable because of the specific updating manner. To improve the efficiency of GPU, we discard this setting and update each position in $F$ independently and concurrently. Furthermore, we store images in batches to take full advantage of the computing units on GPU \cite{kochura2020batch}. The implementation of this algorithm is highly parallel.

\subsection{Image Remapping}
\label{subsection:remapping}

\begin{algorithm}
  \small
  \caption{Memory-efficient image remapping algorithm}
  \label{algorithm:memory_efficient_image_remapping}
  \begin{algorithmic}
    \Statex \textbf{Input:} $S\in \mathbb{R}^{h\times w\times 3}$: source image
    \Statex \textbf{Input:} $F\in \mathbb{N}^{h\times w\times 2}$: estimated NNF
    \Statex \textbf{Input:} $(x,y)$: coordinate of the pixel to be computed
    \State $\hat T(x,y)\gets \boldsymbol{0}\in\mathbb{R}^3$
    \For{$d_x=-p,\cdots,p$}
      \For{$d_y=-p,\cdots,p$}
        \State $(x',y')\gets F(x+d_x,y+d_y)-(d_x,d_y)$
        \State $\hat T(x,y)\gets \hat T(x,y)+S(x',y')$
      \EndFor
    \EndFor
    \State $\hat T(x,y)\gets \frac{\hat T(x,y)}{(2p+1)^2}$
    \State \Return $\hat T(x,y)$
  \end{algorithmic}
\end{algorithm}

Once we obtain the estimated NNF, we can reconstruct the target image using the source image. First, the source image is converted into $h\times w$ patches with shape of $(2p+1)\times(2p+1)\times 3$. Then, the patches are rearranged according to the NNF $F$. Finally, compute the average at the overlapping parts to obtain the reconstructed target image. Note that the VRAM required for storing the patches is $(2p+1)^2$ times of a single image, making it difficult to implement. To reduce the VRAM required and improve the IO efficiency, we directly compute each pixel in the reconstructed image, instead of storing the intermediate results. The pseudocode of this algorithm is presented in Algorithm \ref{algorithm:memory_efficient_image_remapping}. We successfully reduce the space complexity from $\mathcal O(hwp^2)$ to $\mathcal O(hw)$. This function is compiled using NVCC compiler \cite{grover2012compiling} and runs on NVIDIA GPUs. Similar to Algorithm \ref{algorithm:base_patch_matching}, this algorithm also supports batched data.

\subsection{Blending}
\label{subsection:blending}

The first usage of FastBlend is video deflickering. Recently, We witnessed some prior approaches \cite{ceylan2023pix2video, khachatryan2023text2video} that utilize off-the-shelf diffusion models to synthesize videos without training. Given an $N$-frame video $\{G_i\}_{i=0}^{N-1}$ (called guide video), these methods can render the frames into another style. We use $\{S_i\}_{i=0}^{N-1}$ to represent the rendered frames (called style video). The primary challenge of video synthesis is consistency. The frames synthesized by diffusion models may contain inconsistent content because each frame is processed independently. Leveraging the patch matching and image remapping algorithm, we blend the frames in a sliding window together, i.e. let
\begin{equation}
  \label{equation:loss_function_2}
  \tilde{S}_i=\frac{1}{2M+1}\sum_{j=i-M}^{i+M} \left(
    S_j \to S_i,
  \right)
\end{equation}
where $(S_j \to S_i)$ is the remapped frame from $S_j$ to $S_i$ using $\text{NNF}(G_j,G_i)$. In this application scenario, the estimated NNF is applied to a frame other than the source image, which sometimes makes the remapped image looks fragmented. Inspired by Ebsynth \cite{jamrivska2019stylizing}, we use an improved loss function.
\begin{equation}
  \begin{aligned}
    \mathcal L(G_j,G_i,F)_{x,y}=&\alpha||G_j[F(x,y)]-G_i[x,y]||_2^2\\
                          &+||S_j[F(x,y)]-\hat S_i[x,y]||_2^2,
  \end{aligned}
\end{equation}
where $\alpha$ is a hyperparameter to determine how much motion information in the input video will be used for remapping. Note that $\hat S_i$ is the remapped frame, which should be also updated when the NNF is updated. But updating $\hat S_i$ is slower than updating NNF because of random VRAM IO operations. We only update $\hat S_i$ once at the beginning of each iteration. Empirically, we find it has almost no influence on image quality but can make the program much faster. After the remapping program, the blended frames $\{\tilde{S}_i\}_{i=0}^{N-1}$ compose a consistent and fluent video.

In some application scenarios, such as the movie industry, the video quality is important. And in other scenarios, the computational efficiency is more important. To meet different requirements, we devise three inference modes for blending. The inference mode without other modifications is named as balanced mode. The other two modes are fast mode and accurate mode, which are designed to improve efficiency and video quality respectively.

\subsubsection{Efficiency Improvement}

When we implement the blending algorithm in a simple way, we need $\mathcal O(NM)$ times NNF estimation. If $M$, the size of the sliding window, is too large, this algorithm will be extremely slow. Recently, Duan et al. \cite{duan2023diffsynth} proposed a fast patch blending algorithm, which has been demonstrated to achieve $\mathcal O(N\log N)$ time complexity. However, it can only blend the frames in the whole video, not in a sliding window. This algorithm may make the video smoggy when the video is long. We modify the fast patch blending algorithm to meet our requirements and improve efficiency. First, we process every image remapping task independently in parallel, making it faster. Second, we propose a new query algorithm to support blending frames in a sliding window.

\begin{figure*}[]
  \centering
   \includegraphics[width=1.0\linewidth]{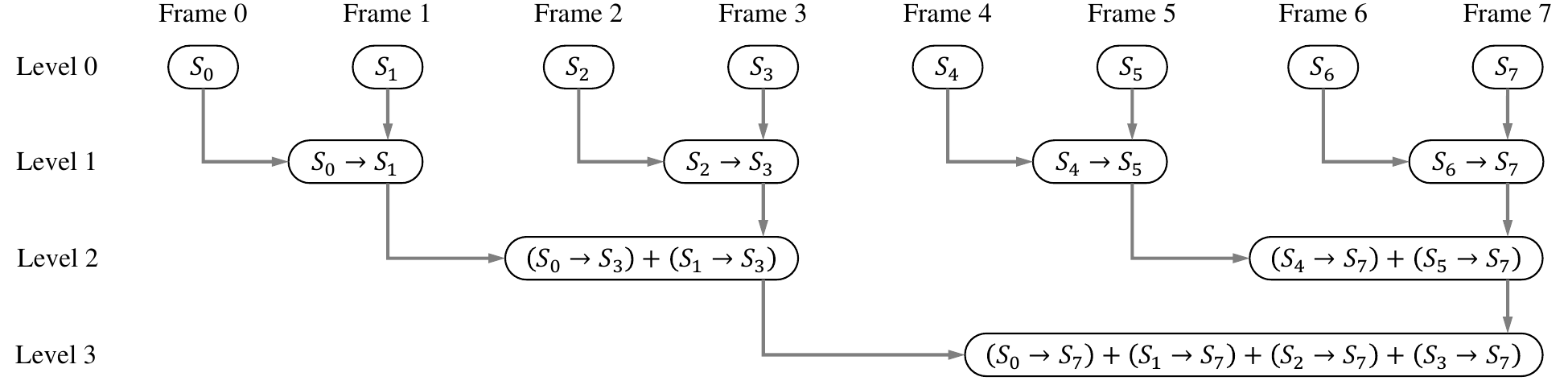}
   \caption{An example of a remapping table in the fast patch blending algorithm.}
   \label{figure:remapping_table}
\end{figure*}

\begin{algorithm}[]
  \small
  \caption{Building a remapping table}
  \label{algorithm:building_a_remapping_table}
  \begin{algorithmic}
    \Statex \textbf{Input:} $\{G_i\}_{i=0}^{N-1}$: guide video
    \Statex \textbf{Input:} $\{S_i\}_{i=0}^{N-1}$: style video
    \State $L_{\text{max}} \gets \lceil\log_2 N\rceil$
    \State Initialize RemappingTable
    \For{$i=0,1,\cdots,N-1$}
      \State $\text{RemappingTable}(i,0)\gets S_i$
      \State $j\gets i$
      \For{$L=0,1,\dots,L_{\text{max}}-1$}
        \If{$\text{BitwiseAnd}(i,2^L)>0$}
          \State continue
        \EndIf
        \State $j\gets \text{BitwiseOr}(j,2^L)$
        \If{$j<N$}
          \State Compute $\text{NNF}(G_i,G_j)$
          \State Compute $(S_i\to S_j)$ using $\text{NNF}(G_i,G_j)$
          \State Add $(S_i\to S_j)$ to $\text{RemappingTable}(j,L+1)$
        \EndIf
      \EndFor
    \EndFor
    \State \Return $\text{RemappingTable}$
  \end{algorithmic}
\end{algorithm}

\begin{algorithm}[]
  \small
  \caption{Building a blending table}
  \label{algorithm:building_a_blending_table}
  \begin{algorithmic}
    \Statex \textbf{Input:} $\{G_i\}_{i=0}^{N-1}$: guide video
    \Statex \textbf{Input:} $\{S_i\}_{i=0}^{N-1}$: style video
    \State Computing $\text{RemappingTable}$ using algorithm \ref{algorithm:building_a_remapping_table}
    \State $L_{\text{max}} \gets \lceil\log_2 N\rceil$
    \State Initialize BlendingTable
    \For{$i=0,1,\cdots,N-1$}
      \State $\text{BlendingTable}(i,0)\gets S_i$
      \For{$L=1,2,\dots,L_{\text{max}}-1$}
        \State $\begin{aligned}\text{BlendingTable}(i,L)\gets &\text{BlendingTable}(i,L-1)\\&+\text{RemappingTable}(i,L)\end{aligned}$
      \EndFor
    \EndFor
    \State \Return $\text{BlendingTable}$
  \end{algorithmic}
\end{algorithm}

\begin{algorithm}[]
  \small
  \caption{Query algorithm on a blending table}
  \label{algorithm:query_algorithm_on_a_blending_table}
  \begin{algorithmic}
    \Statex \textbf{Input:} $\text{BlendingTable}$
    \Statex \textbf{Input:} $l,r$: query interval
    \State Initialize result $A\gets O$
    \State $i\gets r$
    \While{$i\ge l$}
      \State $L\gets 0$
      \While{$\text{BitwiseAnd}(i,2^L)>0$ and $i-2^{L+1}+1\ge l$}
        \State $L\gets L+1$
      \EndWhile
      \State Compute $\text{NNF}(G_i,G_r)$
      \State Compute $(\text{BlendingTable}(i,L)\to S_j)$ using $\text{NNF}(G_i,G_r)$
      \State $A\gets A+(\text{BlendingTable}(i,L)\to S_j)$
    \EndWhile
    \State $A\gets \frac{A}{r-l+1}$
    \State \Return $A$
  \end{algorithmic}
\end{algorithm}

\begin{figure*}[]
  \centering
   \includegraphics[width=1.0\linewidth]{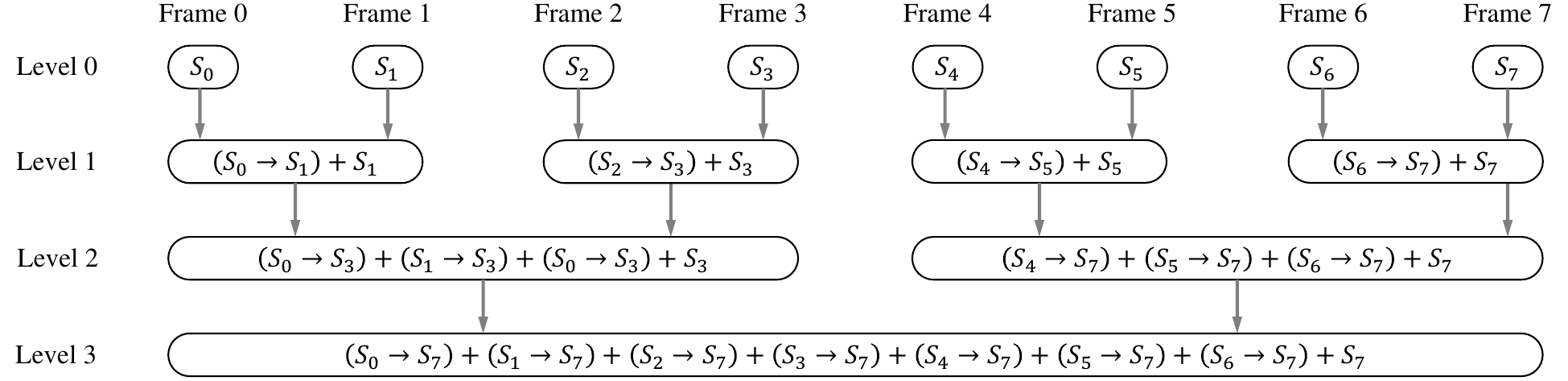}
   \caption{An example of a blending table in the fast patch blending algorithm.}
   \label{figure:blending_table}
\end{figure*}

In the fast patch blending algorithm, the frames are arranged in a remapping table, which is a tree-like array. An example of a remapping table is presented in Figure \ref{figure:remapping_table}. This data structure is similar to some tree-like arrays \cite{fenwick1994new, de2000computational}. Since we have
\begin{equation}
  (S_i\to S_j)\to S_k\doteq S_i\to S_k,
\end{equation}
\begin{equation}
  (S_i+S_j)\to S_k\doteq (S_i\to S_k) + (S_j\to S_k),
\end{equation}
the remapped and blended frames can also be remapped again. In the original version of this algorithm, the remapping table is constructed within $\mathcal O(N)$ time complexity utilizing this conclusion. Yet, this procedure cannot be fully implemented in parallel. In practice, we find that computing each estimated NNF in parallel is faster, although the time complexity is higher. After $\mathcal O(N\log N)$ times of NNF estimation and image remapping, we obtain the whole remapping table. The pseudocode of constructing a remapping table is presented in Algorithm \ref{algorithm:building_a_remapping_table}. Then we construct a blending table to store the prefix sum of each column. Please see Algorithm \ref{algorithm:building_a_blending_table} and Figure \ref{figure:blending_table} for more details.

After constructing the blending table, we propose a new query algorithm based on this data structure. The pseudocode is in Algorithm \ref{algorithm:query_algorithm_on_a_blending_table}. This algorithm can compute $\frac{1}{r-l+1}\sum_{i=l}^r (S_i\to S_r)$ within $\mathcal O(\log N)$ times of NNF estimation. Similarly, we use another symmetric blending table to compute $\frac{1}{r-l+1}\sum_{i=l}^r (S_i\to S_l)$. The two intervals compose a sliding window.
\begin{equation}
\small
  \sum_{j=i-M}^{i+M} (S_j \to S_i)=\sum_{j=i-M}^{i}(S_j \to S_i) + \sum_{j=i}^{i+M}(S_j \to S_i) - S_i.
\end{equation}
All NNF estimation tasks in Algorithm \ref{algorithm:building_a_remapping_table} and Algorithm \ref{algorithm:query_algorithm_on_a_blending_table} are scheduled by a task manager. In the whole algorithm, we need $\mathcal O(N\log N)$ times of NNF estimation. The overall time complexity is not related to the sliding window size, making it possible to use a large sliding window.

\subsubsection{Quality Improvement}

When the flicker noise in a video is excessively pronounced, blending the frames together usually makes the video smoggy. This pitfall arises due to the lack of consistent remapping of content to identical positions across different frames. To overcome this pitfall, we modify the loss function to align the contents in different frames. When different source images $\{S_i\}_{i=0}^{N-1}$ are remapped to the same target frame $T$, $\{S_i\to T\}_{i=0}^{N-1}$ are supposed to be the same, otherwise the details will be discarded during the average calculation. Therefore, we first compute the average remapped image
\begin{equation}
  \overline T=\frac{1}{N}\sum_{i=0}^{N-1}(S_i\to T).
\end{equation}
Then calculate the distance between $\overline T$ and each remapped image. The modified loss function is formulated as
\begin{equation}
  \label{equation:loss_function_3}
  \begin{aligned}
    \mathcal L(S_i,T,F_i)_{x,y} = 
    &\alpha||S_i[F(x,y)]-T[x,y]||_2^2\\
    &+||S_i[F(x,y)]-\overline T[x,y]||_2^2.
  \end{aligned}
\end{equation}
By minimizing $||S_i[F(x,y)]-\overline T[x,y]||_2^2$, the contents are aligned together. The accurate mode requires $\mathcal O(NM)$ times NNF estimation and image remapping. Additionally, we do not need to construct a blending table and store all frames in RAM. Because the frames are rendered one by one, the space complexity is reduced from $\mathcal O(N)$ to $\mathcal O(M)$. Users can process long videos in accurate mode.

\subsection{Interpolation}
\label{subsection:interpolate}

Another application scenario of FastBlend is video interpolation. To transfer the style of a video, we can render some keyframes and then use FastBlend to render the remaining frames. The keyframes can be rendered by diffusion models or GANs, and can even be painted by human artists. In this workflow, users can focus on fine-tuning the details of keyframes, thus it is easier to create fancy videos. Different from the first scenario, FastBlend doesn't make modifications to keyframes.

\subsubsection{Tracking}
\label{subsubsection:tracking}

Considering remapping a single frame $S$ to several frames $\{T_i\}_{i=0}^{N-1}$ that are continuous in time, we may see high-frequency hopping on the remapped frames $\{S\to T_i\}_{i=0}^{N-1}$ if we compute $\{\text{NNF}(S,T_i)\}_{i=0}^{N-1}$ using loss function (\ref{equation:loss_function_2}) independently. Therefore, we decided to design an object tracking mechanism to keep the remapped frames stable. To achieve this, we use an additional step to expand the updating sequence in Algorithm \ref{algorithm:base_patch_matching}. The $\text{NNF}(S, T_i)$ is similar to $\text{NNF}(S, T_{i+1})$ because the adjacent frames $T_i$ and $T_{i+1}$ are similar, thus we add $T_{i+1}$ and $T_{i-1}$ to the updating sequence of $\text{NNF}(S, T_i)$. The algorithm can leverage the information from $\text{NNF}(S, T_{i-1})$ and $\text{NNF}(S, T_{i+1})$ to calculate $\text{NNF}(S, T_i)$, making the video more fluent. Additionally, the object tracking mechanism can also be applied for blending, but we do not see significant improvement in video quality. We set it to an optional setting in blending.

\subsubsection{Alignment}
\label{subsubsection:alignment}

When the number of keyframes is larger than $2$, the remaining frames $\{\tilde S_i\}_{i=l+1}^{r-1}$ between two rendered keyframes $\{S_l,S_r\}$ are created according to the two keyframes. A naive method is to remap each keyframe and combine them linearly. The combined frame is formulated as
\begin{equation}
  \tilde S_i=\frac{r-i}{r-l}(S_l\to S_i)+\frac{i-l}{r-l}(S_r\to S_i).
\end{equation}
However, the two remapped frames $(S_l\to S_i)$ and $(S_r\to S_i)$ may contain inconsistent contents. We can use the loss function (\ref{equation:loss_function_3}) to align the contents in the two keyframes. To further improve the performance, we designed a fine-grained alignment loss function specially for pairwise remapping.
\begin{equation}
  \label{equation:loss_function_4}
  \begin{aligned}
    \begin{aligned}
      \mathcal L(S_l,T,F_l)_{x,y}=
      &\alpha||S_l[F_l(x,y)]-T[x,y]||_2^2\\
      &+||S_l[F_l(x,y)]-S_r[F_r(x,y)]||_2^2,
    \end{aligned}\\
    \begin{aligned}
      \mathcal L(S_r,T,F_r)_{x,y}=
      &\alpha||S_r[F_r(x,y)]-T[x,y]||_2^2\\
      &+||S_l[F_l(x,y)]-S_r[F_r(x,y)]||_2^2.
    \end{aligned}
  \end{aligned}
\end{equation}
Compared with loss function (\ref{equation:loss_function_3}), the loss function (\ref{equation:loss_function_4}) computes the alignment errors on patches of keyframes, not on the remapped frames. Thus, it is more accurate. This loss function is compatible with the object tracking mechanism and can be enabled at the same time.

\subsection{Implementation Details}

To implement an efficient toolkit, we design the following three kernel functions using C++ and compile these components to make them run on GPU.
\begin{itemize}
  \item \textbf{Remap}. Remapping images in a memory-efficient way, i.e., Algorithm \ref{algorithm:memory_efficient_image_remapping}.
  \item \textbf{Patch error}. Computing the errors of matches, i.e., Algorithm \ref{algorithm:base_patch_matching}. The loss function (\ref{equation:loss_function_2}) and (\ref{equation:loss_function_3}) can be implemented using this kernel function.
  \item \textbf{Pairwise patch error}. Computing the pairwise errors in \ref{subsubsection:alignment} to implement loss function (\ref{equation:loss_function_4}).
\end{itemize}
The other components of FastBlend are implemented using Python for better compatibility. To make FastBlend friendly to creators, we provide a WebUI program for interactive usage. FastBlend can also run as an extension of Stable-Diffusion-WebUI\footnote{\url{https://github.com/AUTOMATIC1111/stable-diffusion-webui}}, which is the most popular UI framework of diffusion models.

\newcommand{\figurewidth}{0.235}
\begin{figure*}[]
\centering
\tabcolsep=3pt
\begin{tabular}{cccc}
\includegraphics[width=\figurewidth\linewidth]{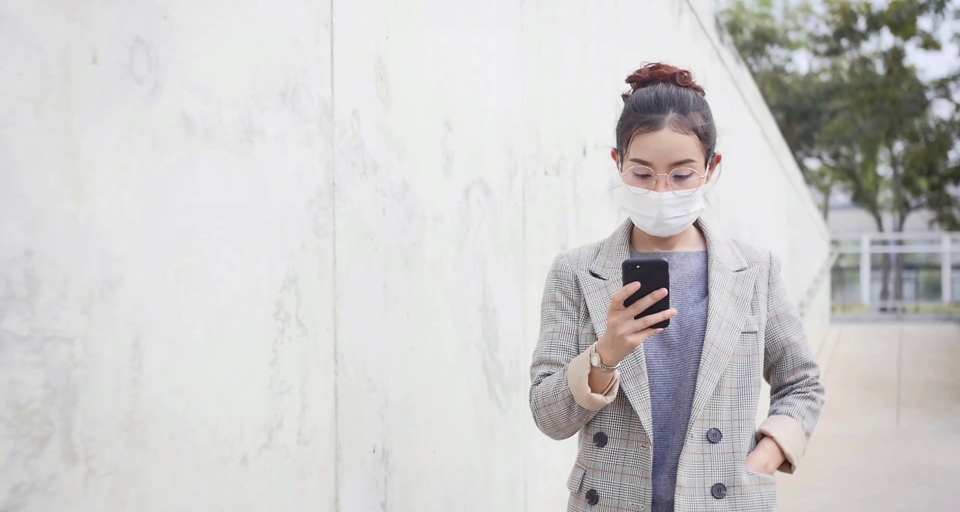} &
\includegraphics[width=\figurewidth\linewidth]{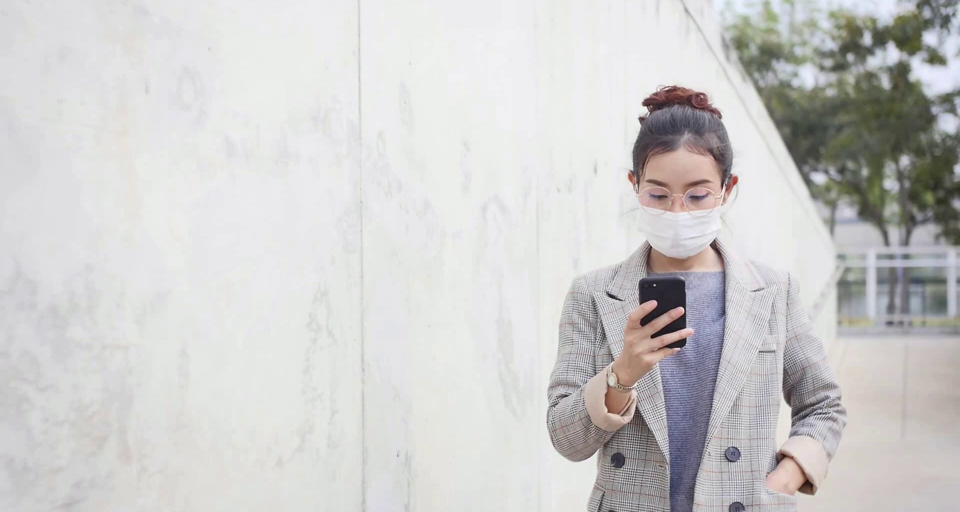} &
\includegraphics[width=\figurewidth\linewidth]{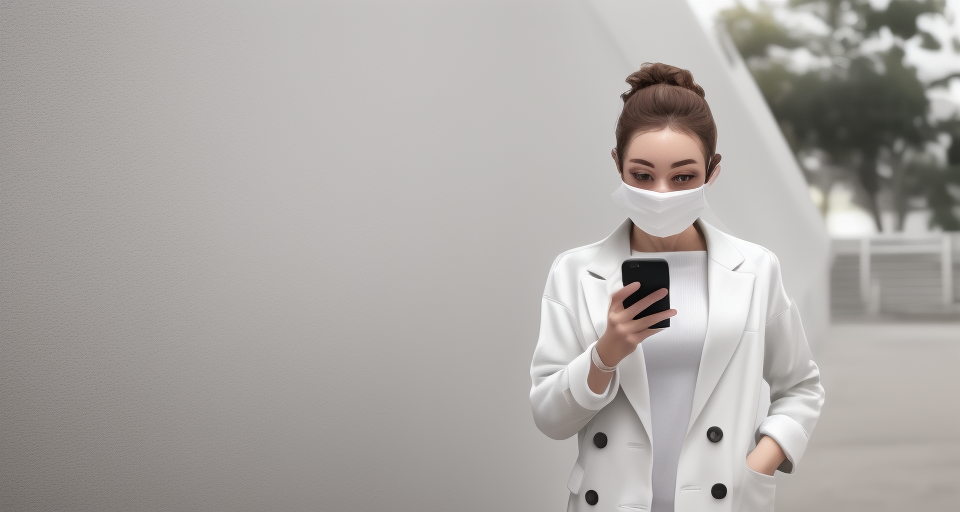} &
\includegraphics[width=\figurewidth\linewidth]{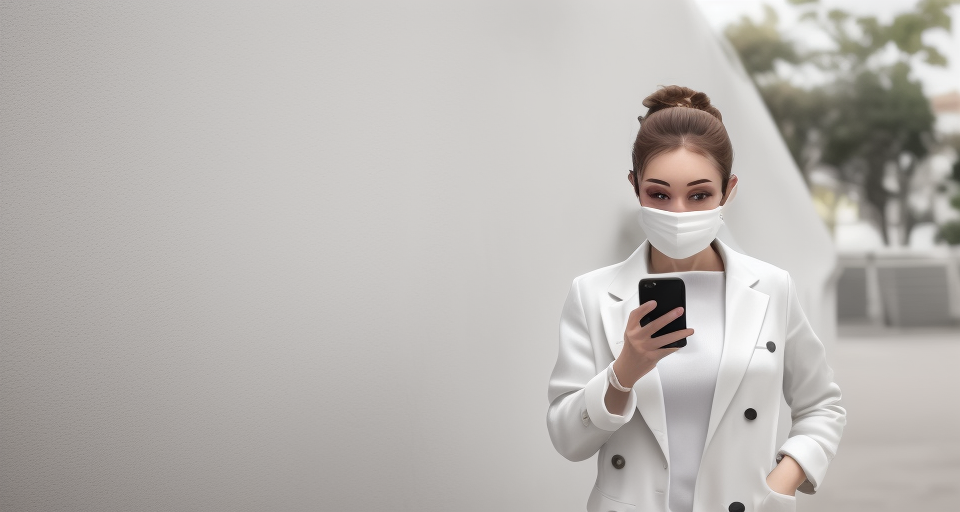} \\
\multicolumn{2}{c}{(a) Input video} & \multicolumn{2}{c}{(b) Naive} \\
\includegraphics[width=\figurewidth\linewidth]{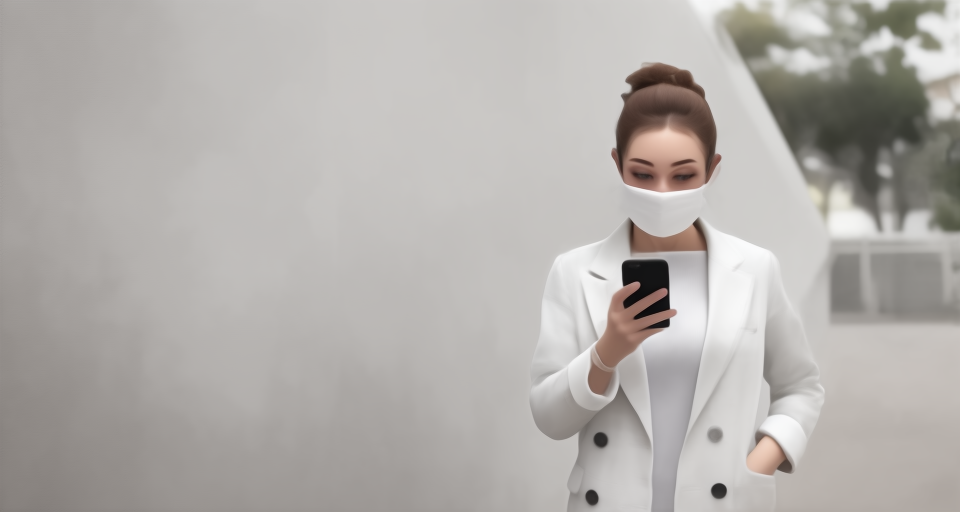} &
\includegraphics[width=\figurewidth\linewidth]{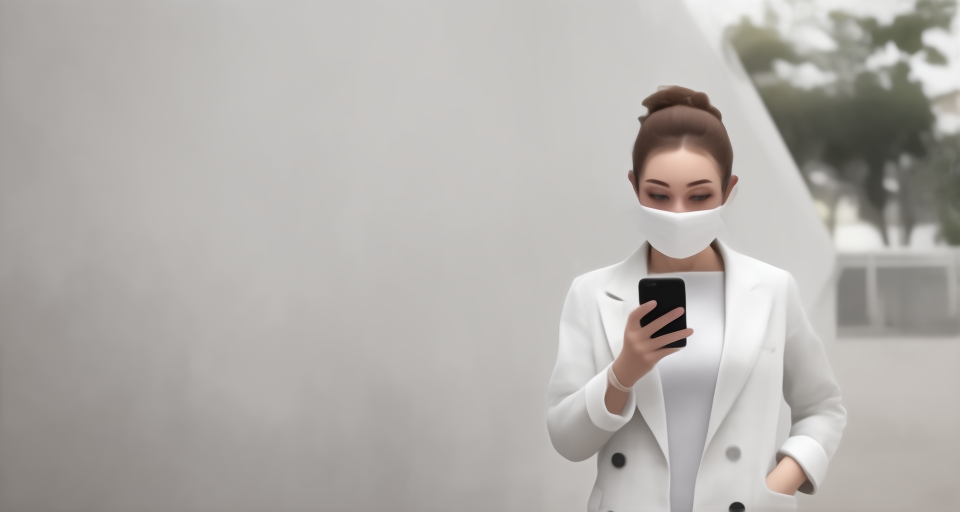} &
\includegraphics[width=\figurewidth\linewidth]{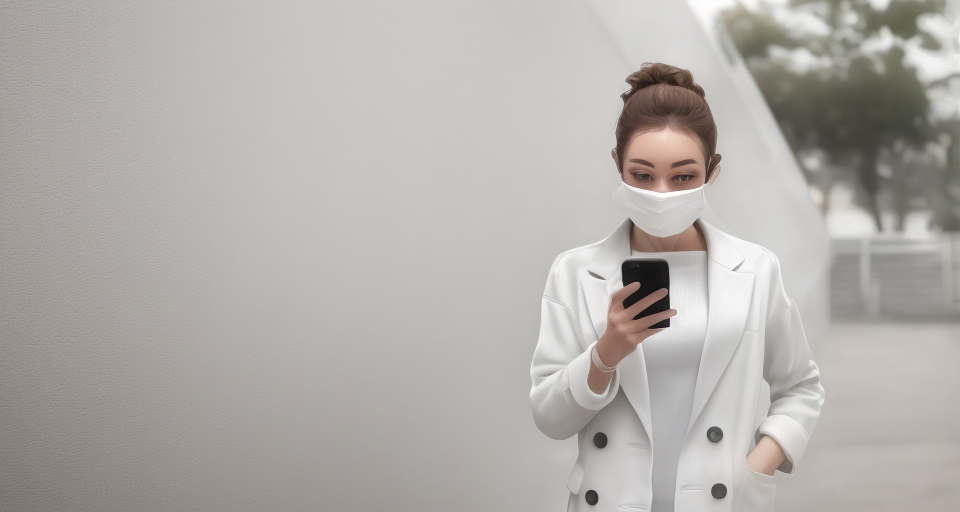} &
\includegraphics[width=\figurewidth\linewidth]{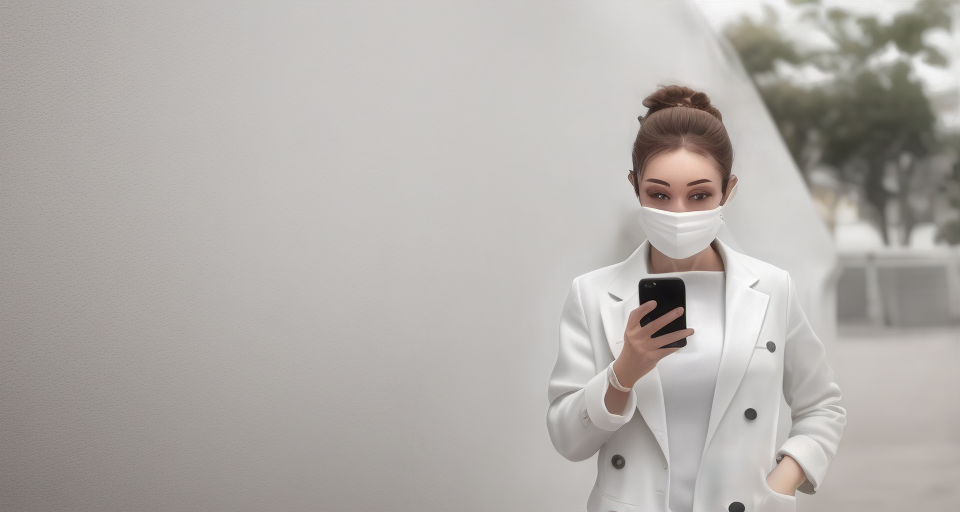} \\
\multicolumn{2}{c}{(c) \textbf{FastBlend}} & \multicolumn{2}{c}{(d) All-In-One Deflicker} \\ 
\includegraphics[width=\figurewidth\linewidth]{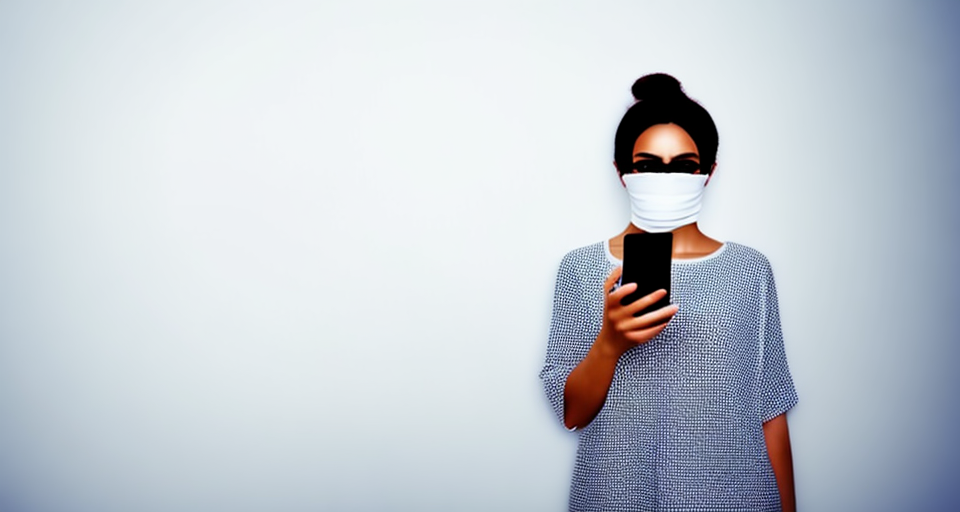} &
\includegraphics[width=\figurewidth\linewidth]{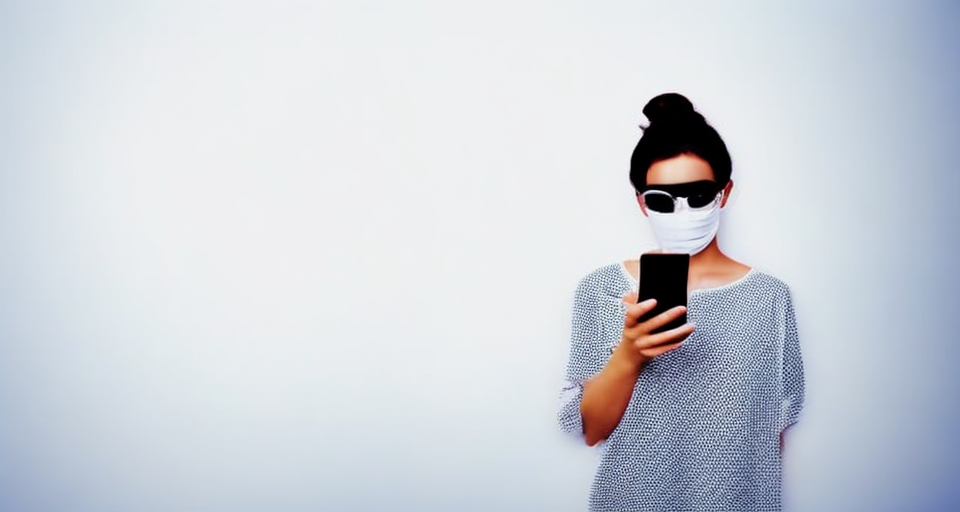} &
\includegraphics[width=\figurewidth\linewidth]{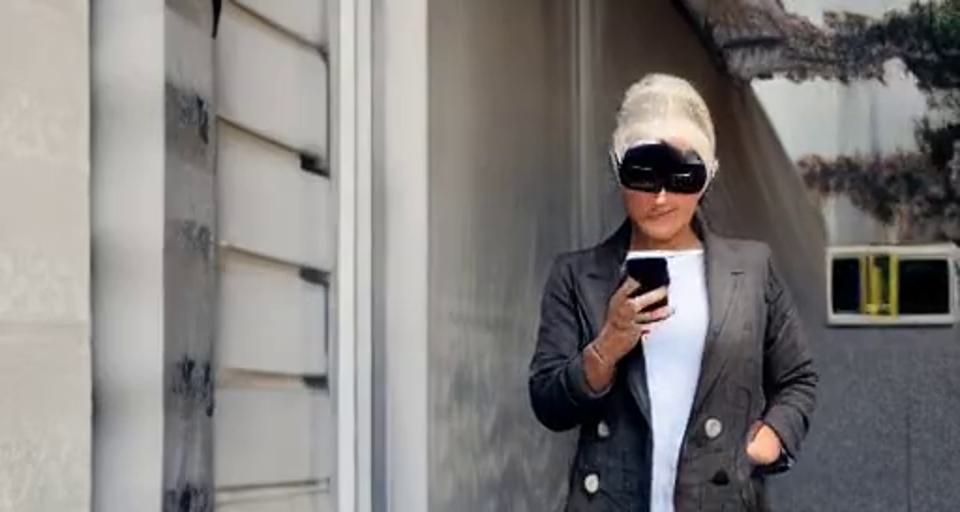} &
\includegraphics[width=\figurewidth\linewidth]{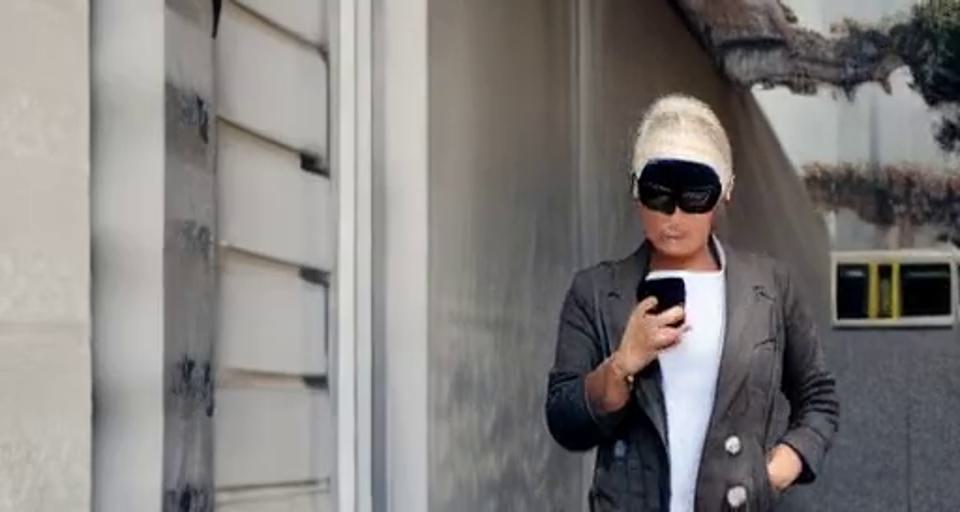} \\
\multicolumn{2}{c}{(e) Pix2Video} & \multicolumn{2}{c}{(f) Text2Video-Zero} \\
\end{tabular}
\caption{Examples in video-to-video translation.}
\label{figure:task1}
\end{figure*}

\begin{figure*}[]
\centering
\tabcolsep=3pt
\begin{tabular}{cccc}
\includegraphics[width=\figurewidth\linewidth]{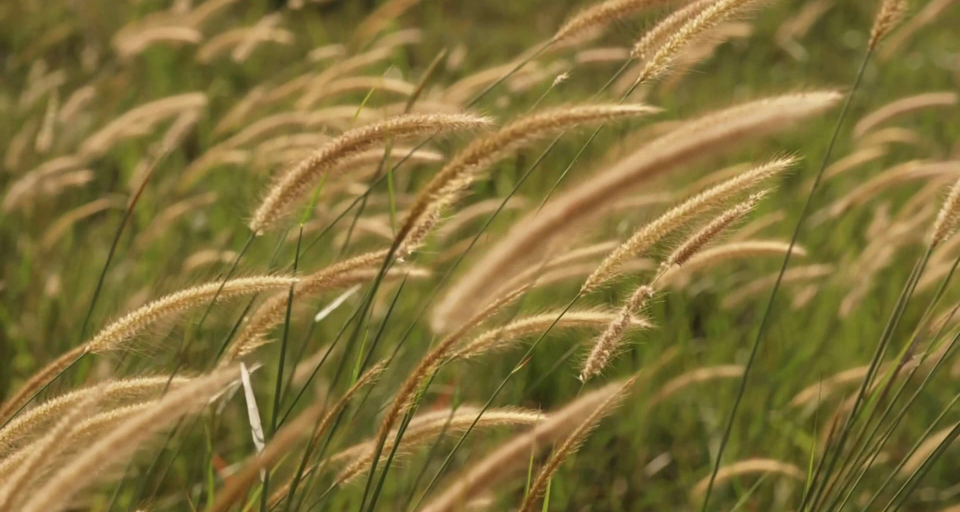} &
\includegraphics[width=\figurewidth\linewidth]{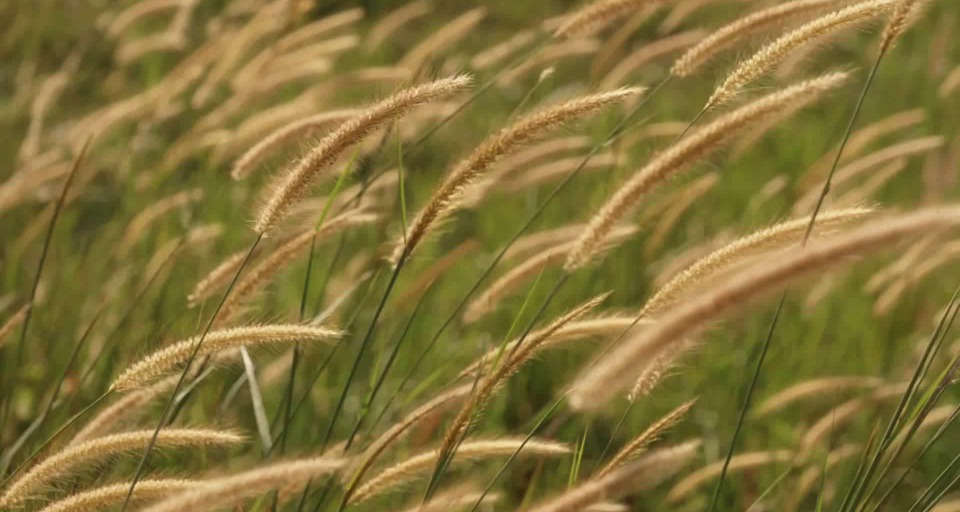} &
\includegraphics[width=\figurewidth\linewidth]{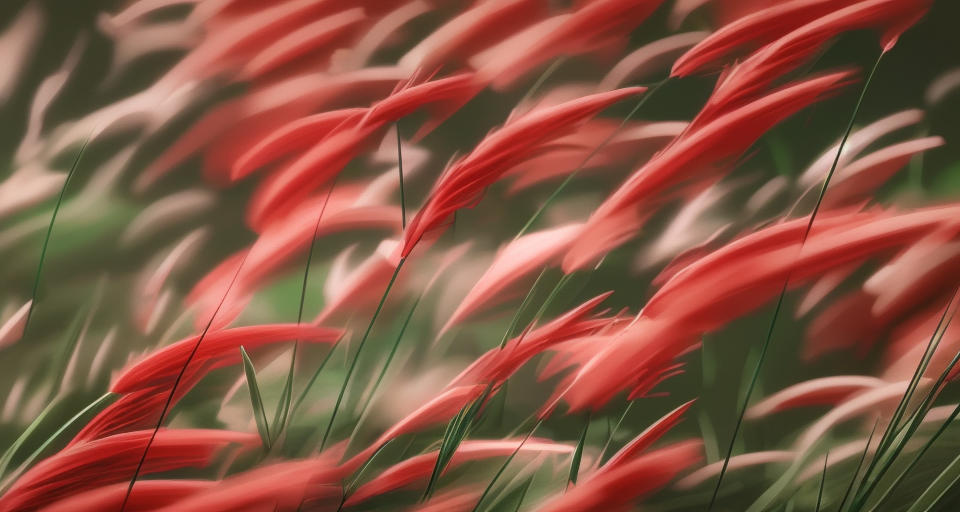} &
\includegraphics[width=\figurewidth\linewidth]{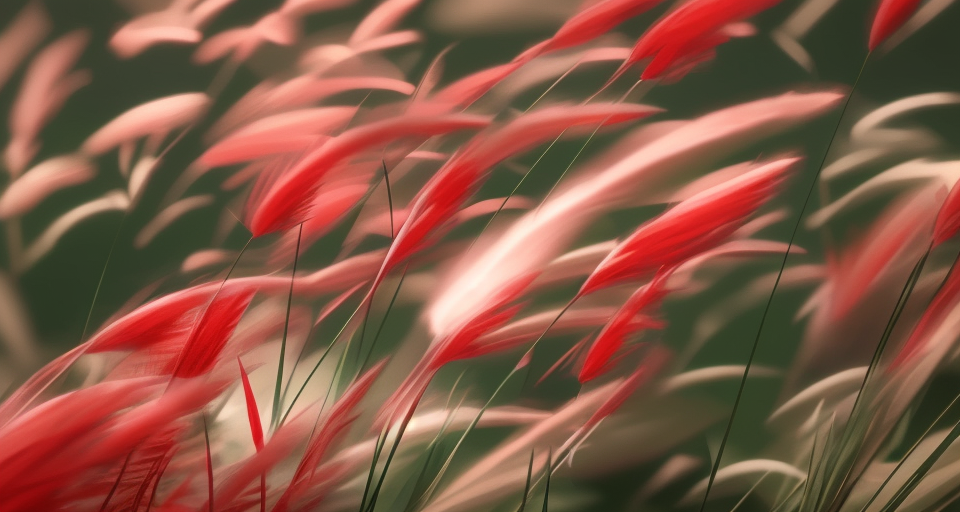} \\
(a) First input frame & (b) Last input frame & (c) First rendered frame & (d) Last rendered frame \\
\includegraphics[width=\figurewidth\linewidth]{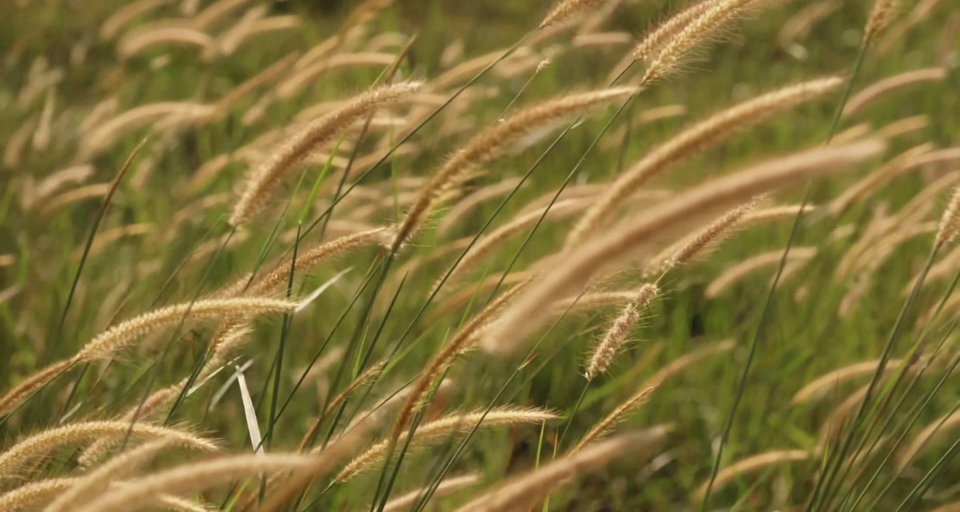} &
\includegraphics[width=\figurewidth\linewidth]{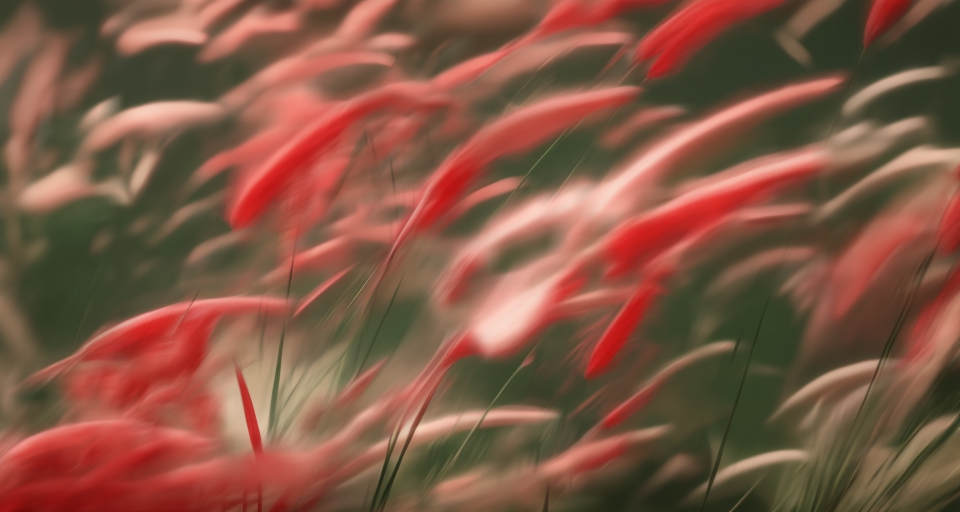} &
\includegraphics[width=\figurewidth\linewidth]{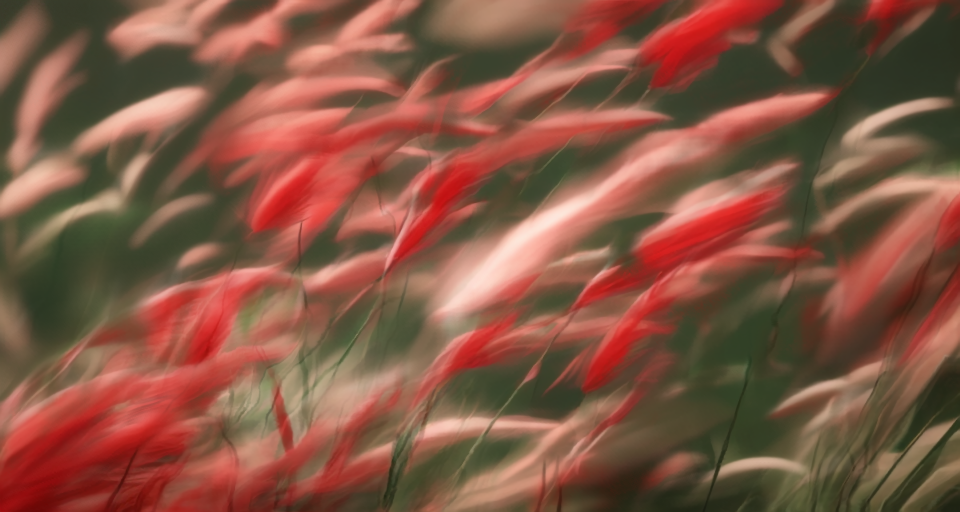} &
\includegraphics[width=\figurewidth\linewidth]{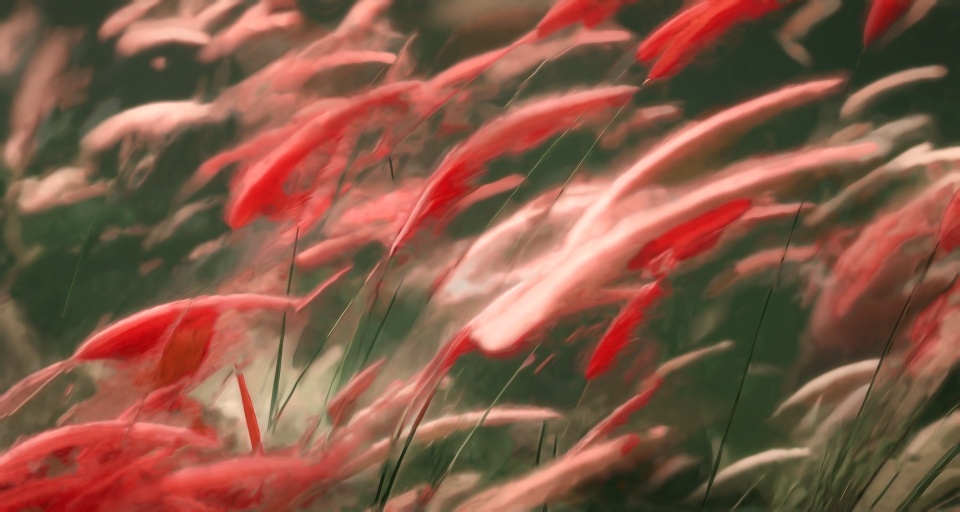} \\
(e) A middle input frame & (f) \textbf{FastBlend} & (g) RIFE & (h) Rerender-A-Video \\
\end{tabular}
\caption{Examples in video interpolation.}
\label{figure:task2}
\end{figure*}

\begin{figure*}[]
\centering
\tabcolsep=3pt
\begin{tabular}{cccc}
\includegraphics[width=\figurewidth\linewidth]{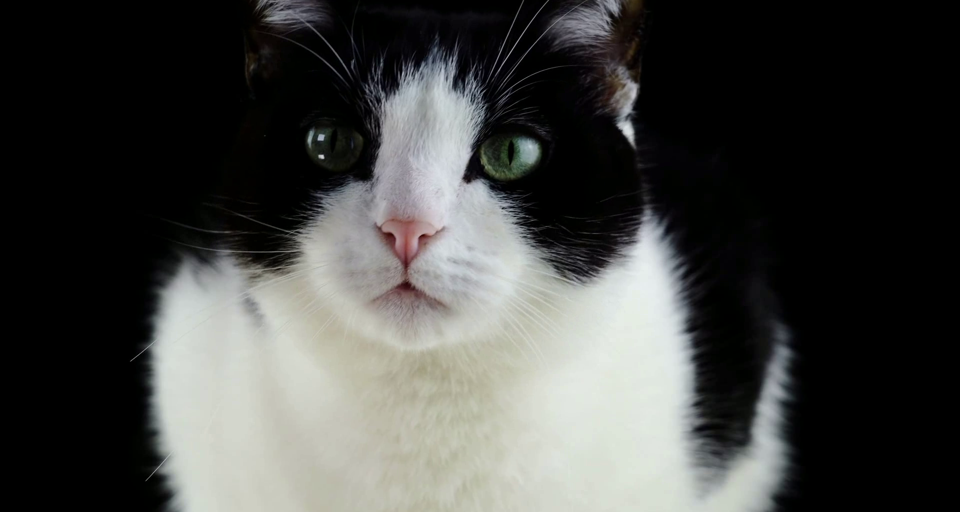} &
\includegraphics[width=\figurewidth\linewidth]{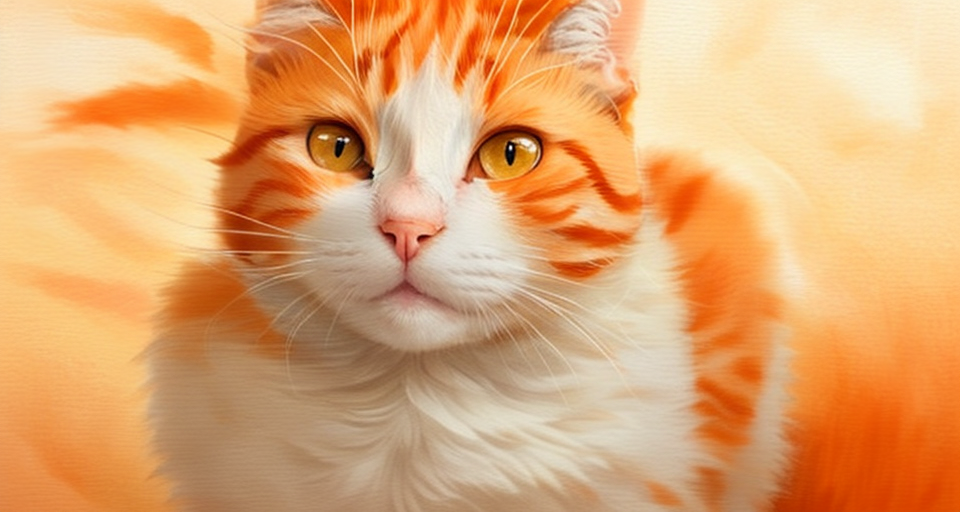} &
\includegraphics[width=\figurewidth\linewidth]{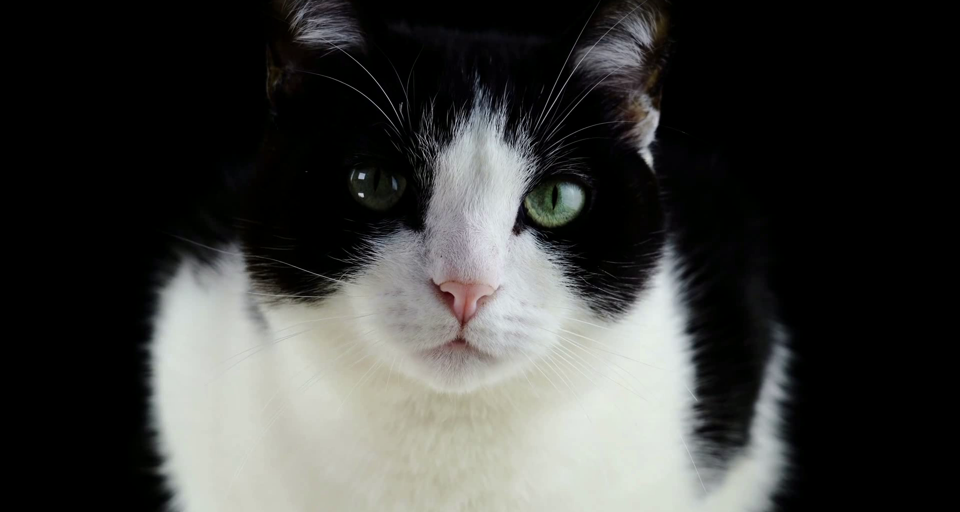} &
\includegraphics[width=\figurewidth\linewidth]{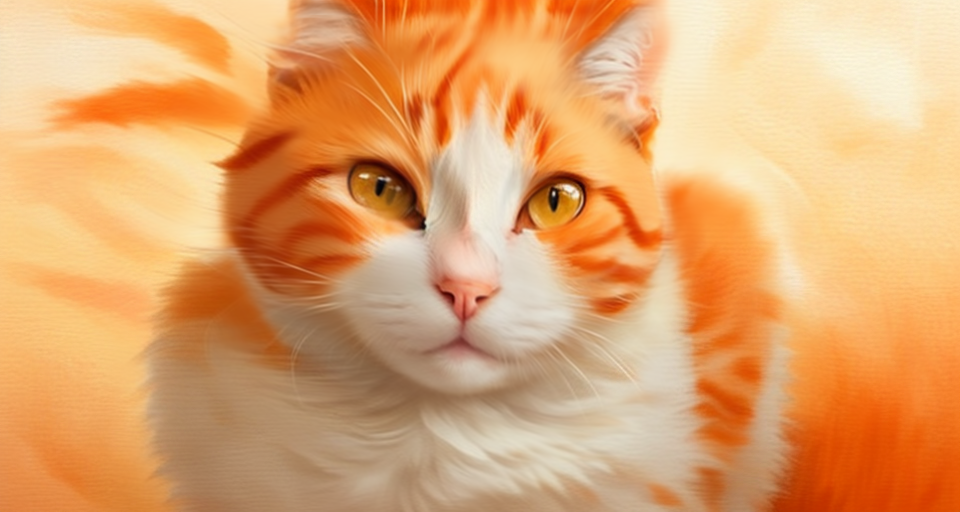} \\
(a) Guide image & (b) Style Image & (c) Input frame & (d) \textbf{FastBlend} \\
\includegraphics[width=\figurewidth\linewidth]{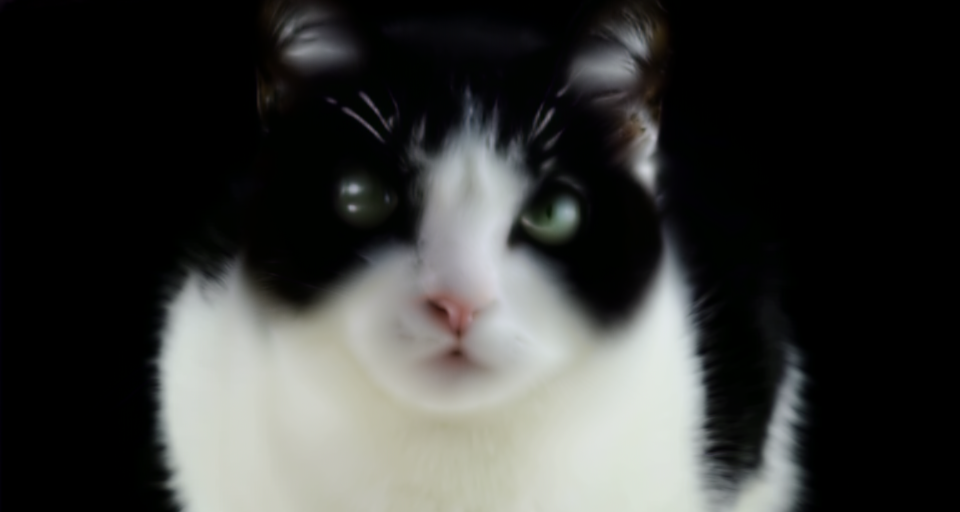} &
\includegraphics[width=\figurewidth\linewidth]{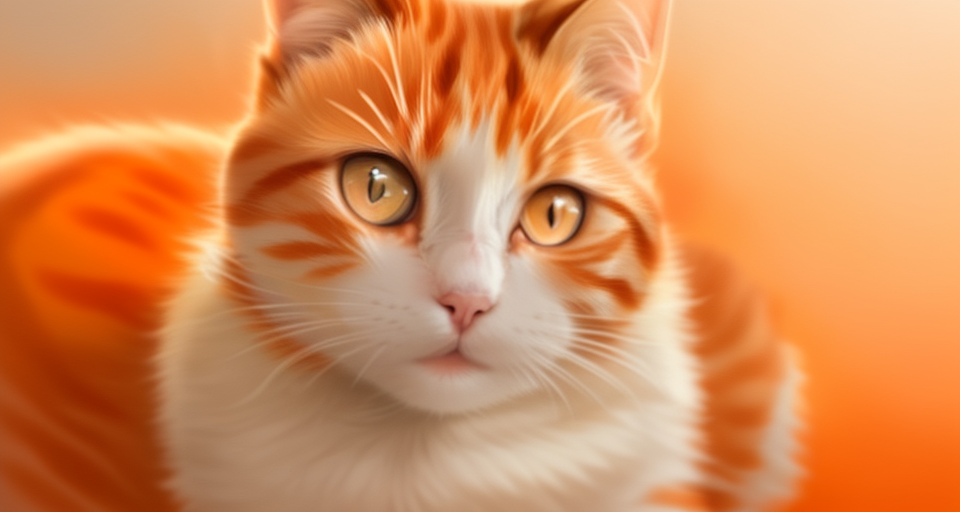} &
\includegraphics[width=\figurewidth\linewidth]{figure/task3/input.png} &
\includegraphics[width=\figurewidth\linewidth]{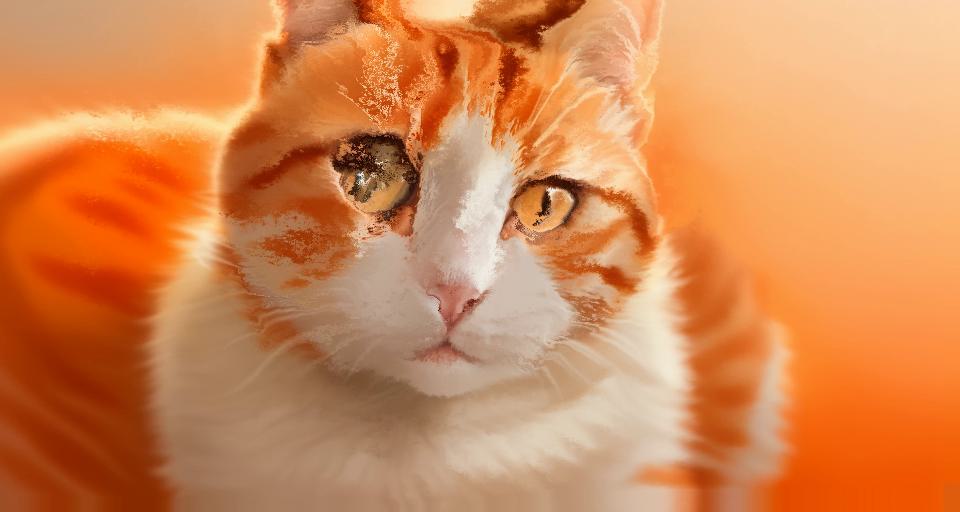} \\
(e) Canonical image & (f) Controlling image & (g) Input frame & (h) CoDeF \\
\end{tabular}
\caption{Examples in image-driven video processing.}
\label{figure:task3}
\end{figure*}

\section{Experiments}

To demonstrate the effectiveness of FastBlend, we evaluate FastBlend and other baseline methods in three tasks. Due to the lack of evaluation metrics \cite{brooks2022generating, ouyang2023codef}, we first present several video samples and then quantify the performance of FastBlend through human evaluation.

\subsection{Video-to-Video Translation}

Given a video and a text prompt, we transfer the style of the video according to the semantic information, while preserving the structural information. In this task, we employ the widely acclaimed diffusion model DreamShaper\footnote{\url{https://civitai.com/models/4384/dreamshaper}} from open-source communities and process the entire video frame by frame. To ensure that the video inherits the original video's structural information, we use two ControlNet \cite{zhang2023adding} models, including SoftEdge and Depth. The number of sampling steps is 20, the ControlNet scale is set to 1.0, the classifier-free guidance scale \cite{ho2021classifier} is 7.5, and the sampling scheduler is DDIM \cite{song2020denoising}. These hyperparameters are tuned empirically. Additionally, we enable the cross-frame attention mechanism to enhance video consistency, which is a widely proven effective trick \cite{yang2023rerender, duan2023diffsynth, qi2023fatezero, ceylan2023pix2video, khachatryan2023text2video}. After that, we perform post-processing on the videos using the fast blending mode in FastBlend to generate fluent videos.

An example video is shown in Figure \ref{figure:task1}, where the prompt is ``a woman, mask, phone, white clothes". Using the aforementioned diffusion model, the input video (Figure \ref{figure:task1}a) is naively processed frame by frame, and the clothes in the video turn white (Figure \ref{figure:task1}b). However, there is a noticeable inconsistency in the position of the buttons. We first compared FastBlend with All-In-One Deflicker \cite{lei2023blind}, which is a state-of-the-art deflickering method. In the results of FastBlend (Figure \ref{figure:task1}c), the positions of the buttons are aligned, and the content in the two frames is consistent. In contrast, All-In-One Deflicker can only eliminate slight flickering and cannot significantly improve video consistency. Furthermore, we compared this pipeline with other video processing algorithms based on diffusion models, including Pix2Video \cite{ceylan2023pix2video} and Text2Video-Zero \cite{khachatryan2023text2video}. The results of these two baseline methods (Figure \ref{figure:task1}e and Figure \ref{figure:task1}f) also exhibit inconsistent content, such as the arms in Figure \ref{figure:task1}e and the face in Figure \ref{figure:task1}f. In this example, we demonstrate that FastBlend, combined with the diffusion model, can generate coherent and realistic videos in video-to-video translation.

\subsection{Video Interpolation}

Given a video, we use the diffusion model to render the first frame and the last frame, and then interpolate the content between these two frames. We employ the same diffusion model and hyperparameters as in the video-to-video translation task to render these two frames, and then use the interpolation mode in FastBlend to generate the video. We also enable the tracking mechanism from section \ref{subsubsection:tracking} and Alignment from section \ref{subsubsection:tracking} to enhance the coherency of the video.

An example video is shown in Figure \ref{figure:task2}, where the original video's first frame (Figure \ref{figure:task2}a) and last frame (Figure \ref{figure:task2}b) are rendered through the diffusion model (Figure \ref{figure:task2}c and Figure \ref{figure:task2}d), respectively. We selected a middle frame from the original video (Figure \ref{figure:task2}e) for comparison of FastBlend and other methods. Due to the significant differences between the two rendered frames, this interpolation task is quite challenging. FastBlend successfully constructs a realistic image (Figure \ref{figure:task2}f). We compare FastBlend to RIFE \cite{huang2022real}, which is a video interpolation algorithm. We observe that RIFE could not generate the swaying grass, and some generated grass is even broken (Figure \ref{figure:task2}g), as it cannot fully utilize the information in the input frame (Figure \ref{figure:task2}e). Rerender-A-Video \cite{yang2023rerender} also includes an interpolation algorithm based on patch matching, but it results in ghosting in the generated video (Figure \ref{figure:task2}h). This example demonstrates the effectiveness of our approach in video interpolation.

\subsection{Image-Driven Video Processing}

Note that the interpolation mode of FastBlend can render an entire video using only one frame, which allows users to finely adjust a single image and then extend that image's style to the entire video. We adopted the diffusion model and experimental settings from the video-to-video translation task and validated FastBlend's performance on this particular task.

We compare the performance of FastBlend with CoDeF \cite{ouyang2023codef}, and an example is shown in Figure \ref{figure:task3}. For FastBlend, we only process the middle frame (Figure \ref{figure:task3}a) into a style image (Figure \ref{figure:task3}b) and then use the interpolation mode to generate the video. Alignment from \ref{subsubsection:alignment} is disabled since there is no need to align information from multiple frames. For CoDeF, the approach starts by generating a canonical image (Figure \ref{figure:task3}e), which is then processed using the diffusion model to create a controlling image (Figure \ref{figure:task3}f) for reference. We selected a frame (Figure \ref{figure:task3}c and Figure \ref{figure:task3}g) from the original video, and the result from FastBlend (Figure \ref{figure:task3}d) is noticeably more realistic compared to CoDeF (Figure \ref{figure:task3}h), even though CoDeF's canonical image has a closer structural resemblance to this frame.

\subsection{Human Evaluation}

To quantitatively evaluate FastBlend and other baseline methods in the three tasks, we conduct comparative experiments on a publicly available dataset. The dataset we used is Pixabay100 \cite{duan2023diffsynth}, which contains 100 videos collected from Pixabay\footnote{https://pixabay.com/videos/} along with manually written prompts. In each task, we process each video with each method according to the experimental settings mentioned above. In video interpolation, it is not feasible to apply RIFE to long videos, thus we only use the first 60 frames of each video. We invite 15 participants for a double-blind evaluation. In each evaluation round, we randomly select one of the tasks, then randomly choose a video and present the participants with videos generated by two different methods. One video is generated by FastBlend, and the other is generated by a randomly selected baseline method. The positions of these two videos were also randomized. Participants are asked to choose the video that looks best in terms of consistency and clarity or choose ``tie'' if the participant cannot determine which video is better.

\begin{table}[]
\caption{The results of human evaluation. Task 1 denotes video-to-video translation, task 2 denotes video interpolation and task 3 denotes image-driven video processing.}
\label{table:human_evaluation}
\footnotesize
\centering
\begin{tabular}{l|l|ccc}
\hline
\multirow{2}{*}{}       & \multirow{2}{*}{Baseline} & \multicolumn{3}{c}{Which one is better} \\ \cline{3-5} 
                        &                           & FastBlend    & Tie        & Baseline    \\ \hline
\multirow{3}{*}{Task 1} & All-In-One Deflicker      & 75.64\%      & 14.10\%    & 10.26\%     \\
                        & Pix2Video                 & 91.49\%      & 4.26\%     & 4.26\%      \\
                        & Text2Video-Zero           & 89.44\%      & 6.83\%     & 3.73\%      \\ \hline
\multirow{2}{*}{Task 2} & RIFE                      & 36.82\%      & 31.82\%    & 31.36\%     \\
                        & Rerender-A-Video          & 32.61\%      & 37.83\%    & 29.57\%     \\ \hline
Task 3                  & CoDeF                     & 64.67\%      & 16.67\%    & 18.67\%     \\ \hline
\end{tabular}
\end{table}

The results of the human evaluation are shown in Table \ref{table:human_evaluation}. In the video-to-video translation and image-driven video processing, the participants unanimously found that FastBlend's overall performance is significantly better than the baseline methods. In the video interpolation, due to the relatively short video length, participants had difficulty observing significant differences, so FastBlend is slightly better than the baseline methods.

\subsection{Efficiency Analysis}

\begin{table}[]
\caption{The time consumed for processing 100 frames.}
\centering
\footnotesize
\label{table:time}
\begin{tabular}{l|l|c}
\hline
                        & Method               & Time consumed          \\ \hline
\multirow{2}{*}{Task 1} & All-In-One Deflicker & 5.42 mins              \\
                        & FastBlend            & \textbf{2.27 mins}     \\ \hline
\multirow{3}{*}{Task 2} & RIFE                 & \textbf{0.09 mins}     \\
                        & Rerender-A-Video     & 3.27 mins              \\
                        & FastBlend            & 1.15 mins              \\ \hline
\multirow{2}{*}{Task 3} & CoDeF                & 4.16 mins              \\
                        & FastBlend            & \textbf{0.67 mins}     \\ \hline
\end{tabular}
\end{table}

FastBlend can fully leverage GPU units. We compare the computational efficiency of these methods using an NVIDIA RTX 4090 GPU and record the time consumed for rendering 100 frames. The experimental results are shown in Table \ref{table:time}. Considering that Pix2Video and Text2Video-Zero are video processing methods strongly coupled with diffusion models, comparing them with other methods is unfair. In video-to-video translation and image-driven video processing tasks, FastBlend outperforms other methods significantly. In video interpolation tasks, FastBlend requires less time than Rerender-A-Video, even though both are designed based on the patch match algorithm. The video interpolation method RIFE is faster than FastBlend, but it only generates gradual effects between keyframes and cannot produce realistic motion effects. Due to the extremely high GPU utilization, FastBlend achieves both high video quality and high efficiency.

\subsection{Comparison of inference modes for blending}

\newcommand{\figurewidthnine}{0.092}
\newcommand{\figurewidththree}{0.3}
\begin{figure*}[]
\centering
\tabcolsep=3pt
\begin{tabular}{ccccccccccc}
... &
\includegraphics[width=\figurewidthnine\linewidth]{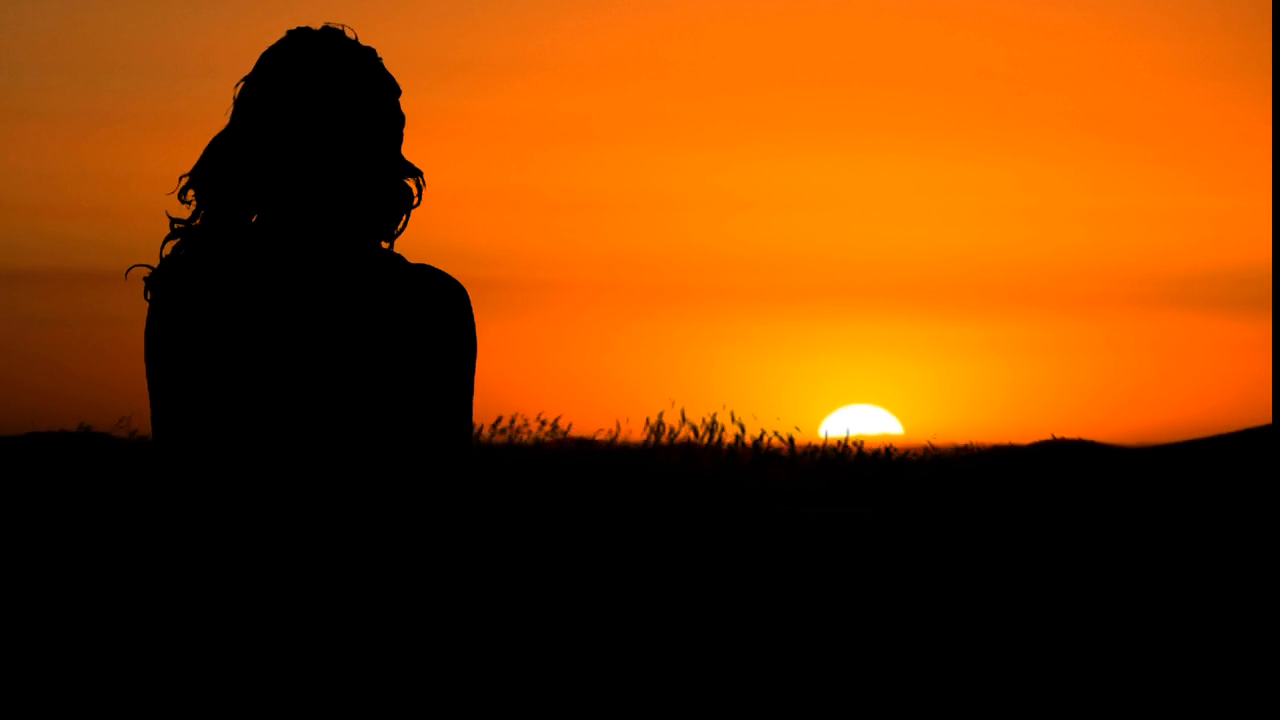} &
\includegraphics[width=\figurewidthnine\linewidth]{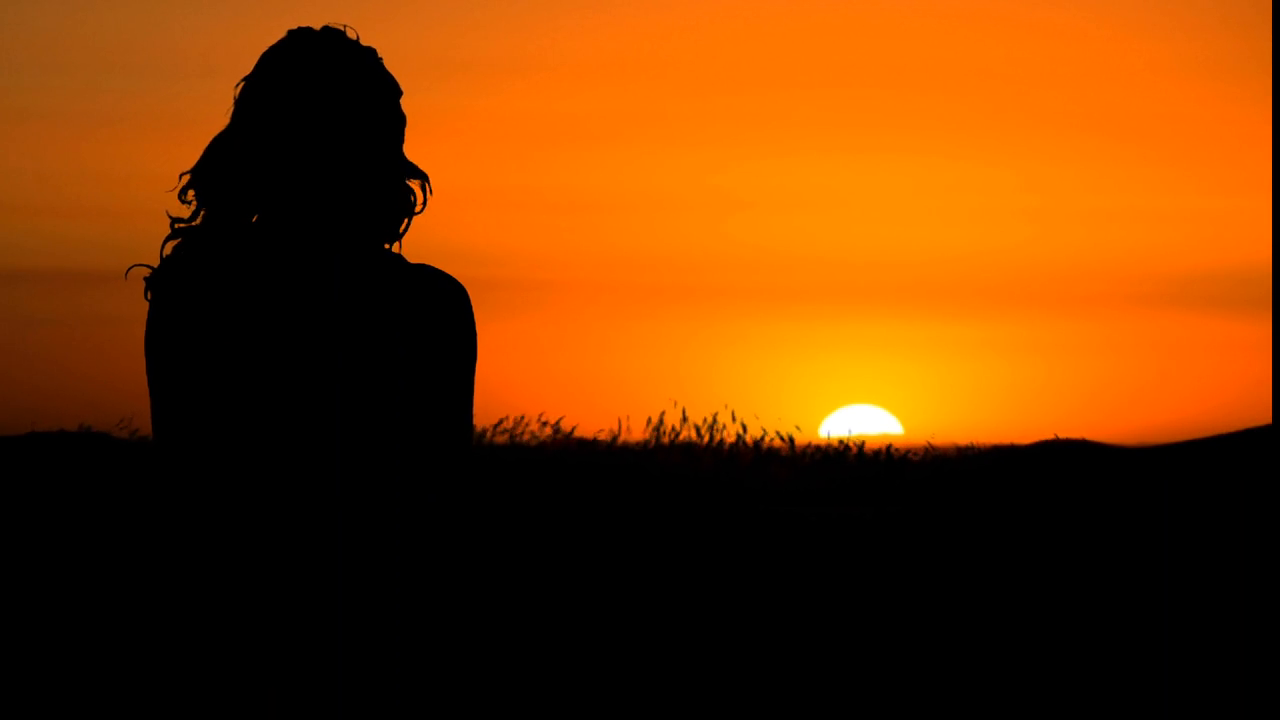} &
\includegraphics[width=\figurewidthnine\linewidth]{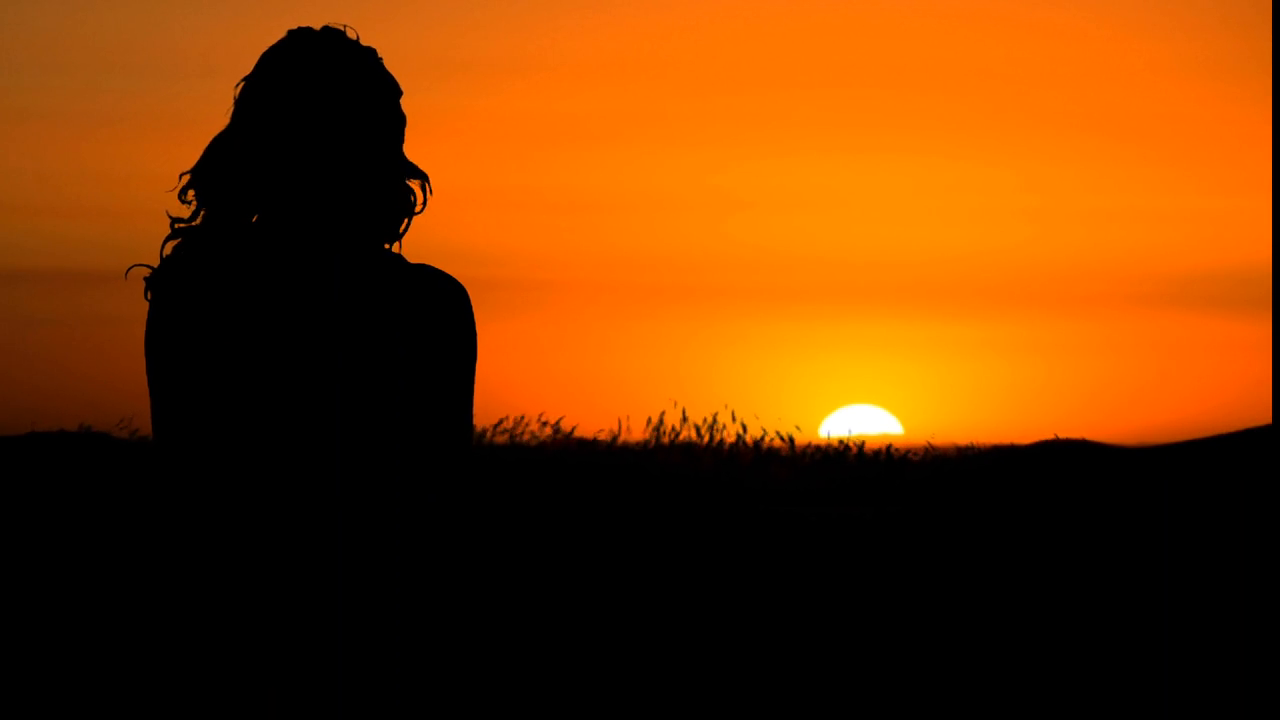} &
\includegraphics[width=\figurewidthnine\linewidth]{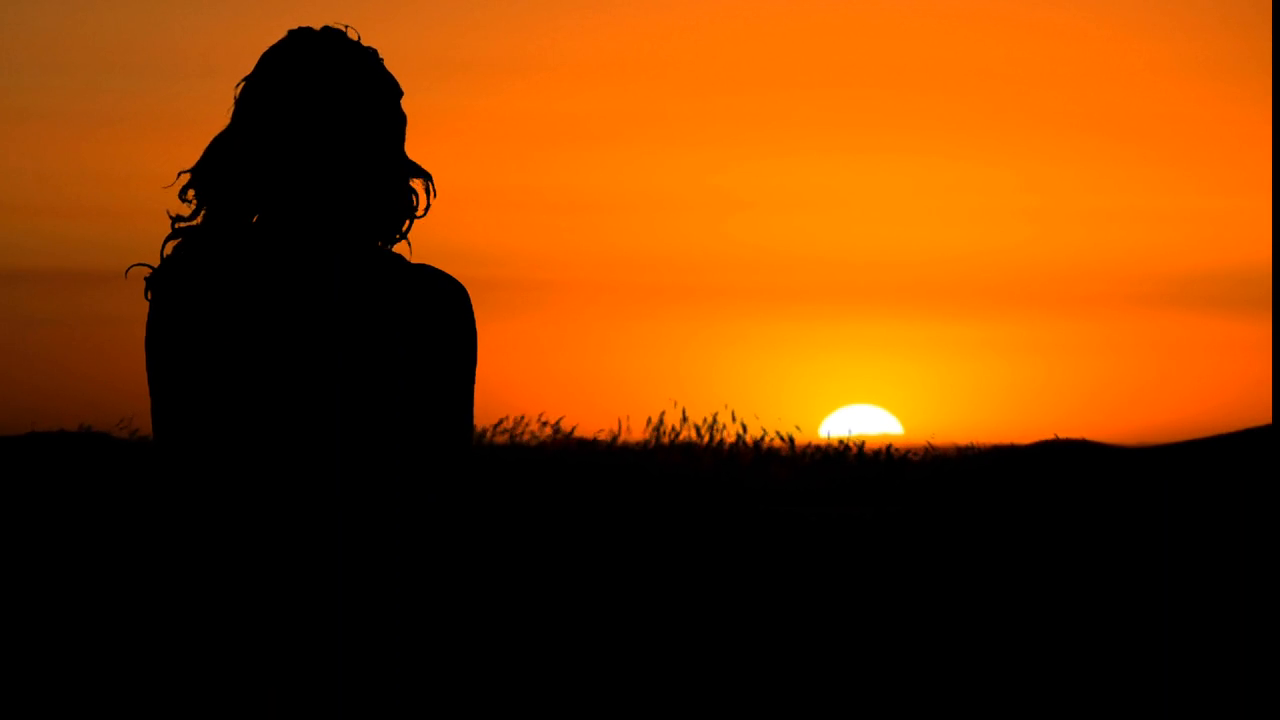} &
\includegraphics[width=\figurewidthnine\linewidth]{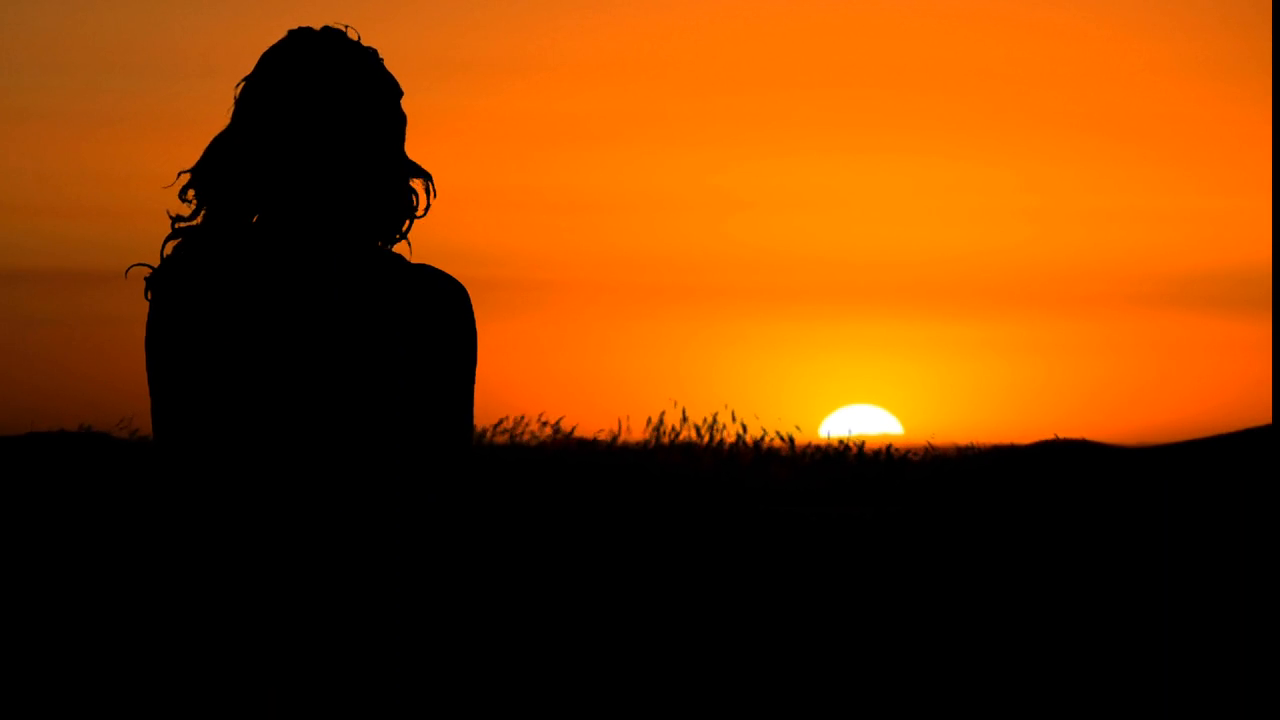} &
\includegraphics[width=\figurewidthnine\linewidth]{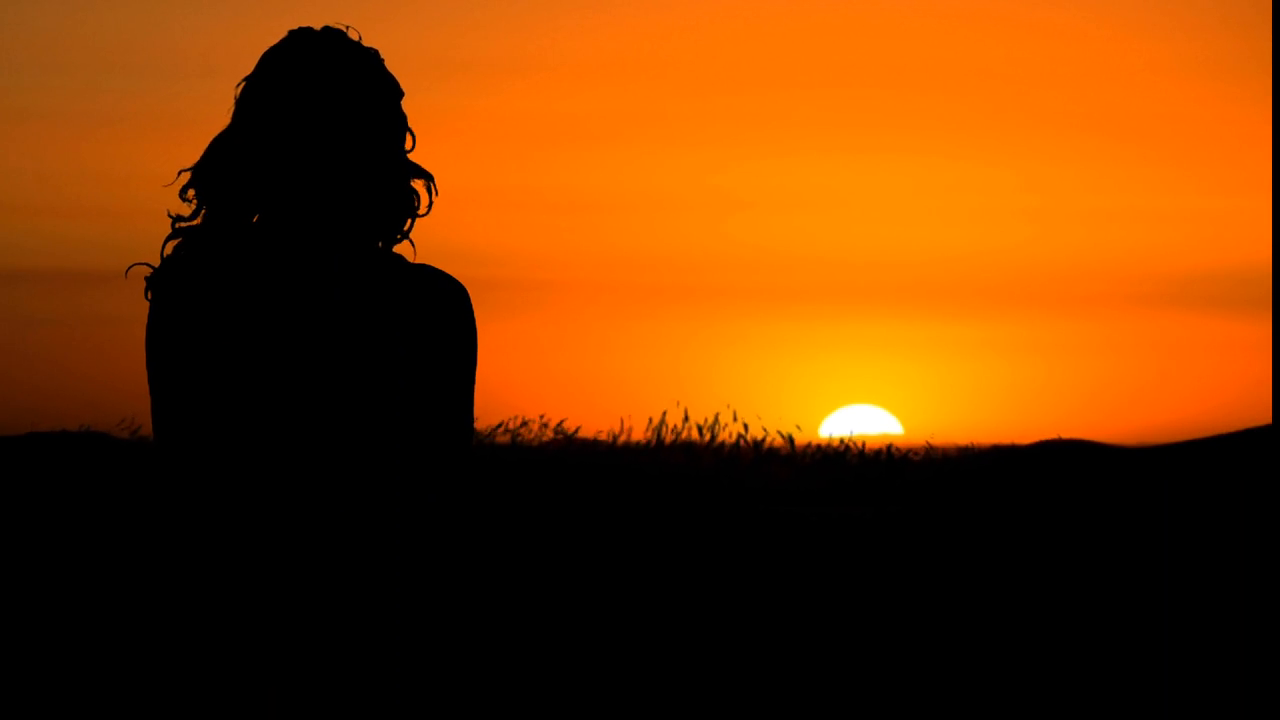} &
\includegraphics[width=\figurewidthnine\linewidth]{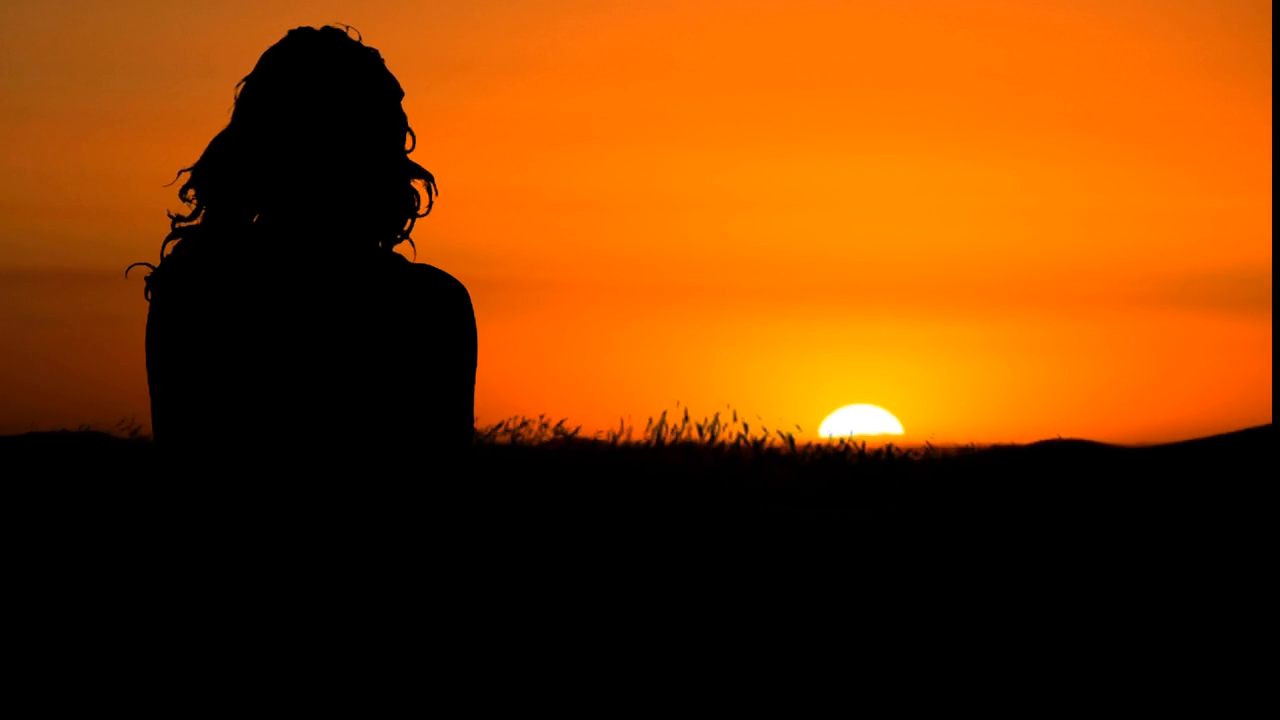} &
\includegraphics[width=\figurewidthnine\linewidth]{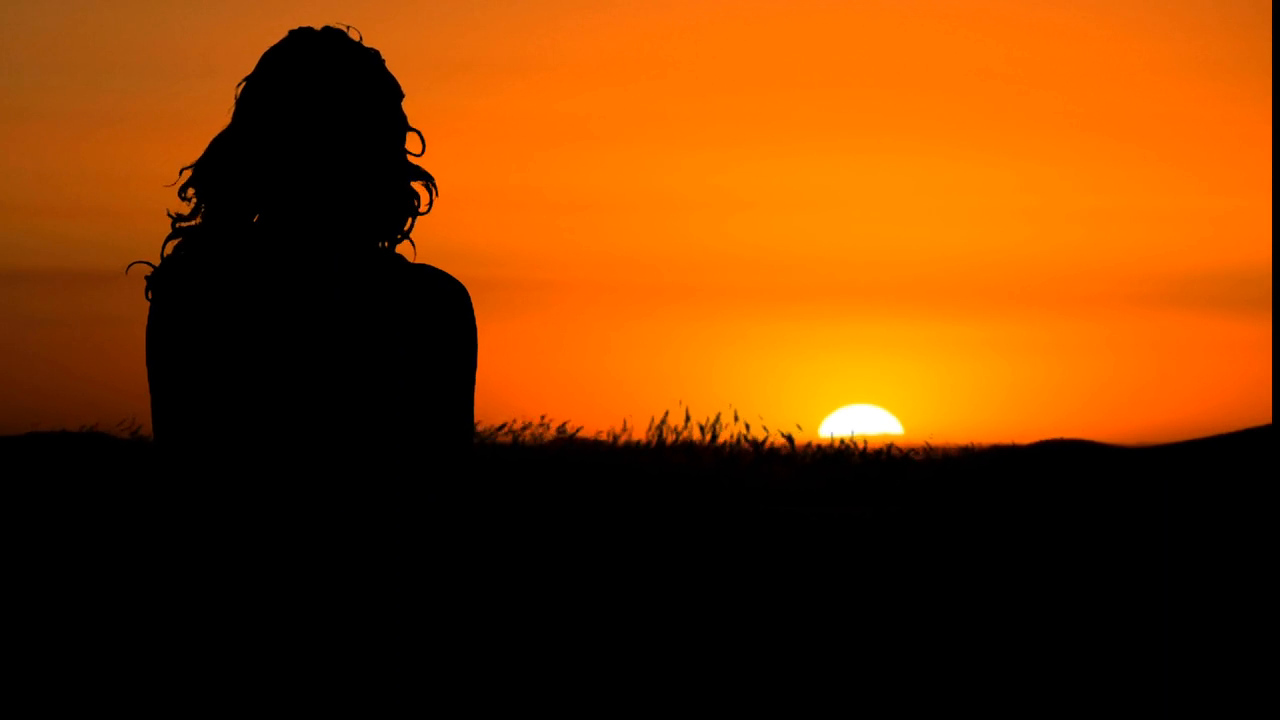} &
\includegraphics[width=\figurewidthnine\linewidth]{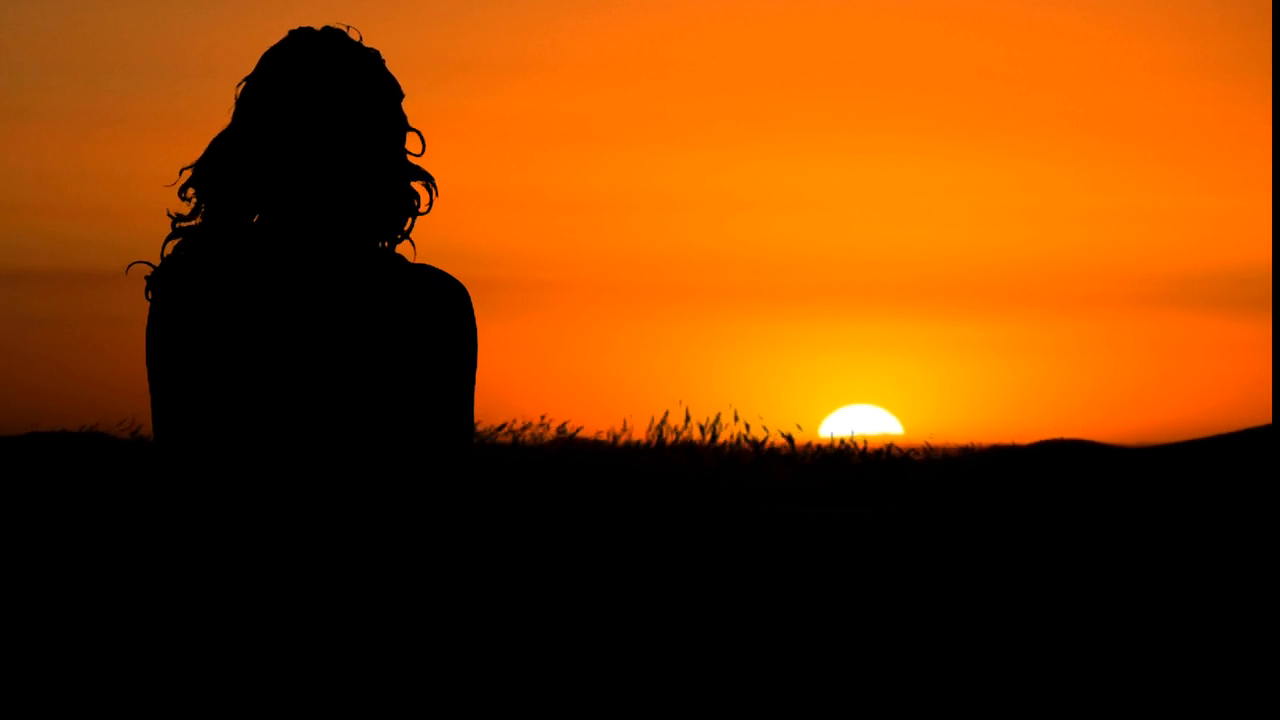} &
... \\
\multicolumn{11}{c}{(a) Original video} \\
... &
\includegraphics[width=\figurewidthnine\linewidth]{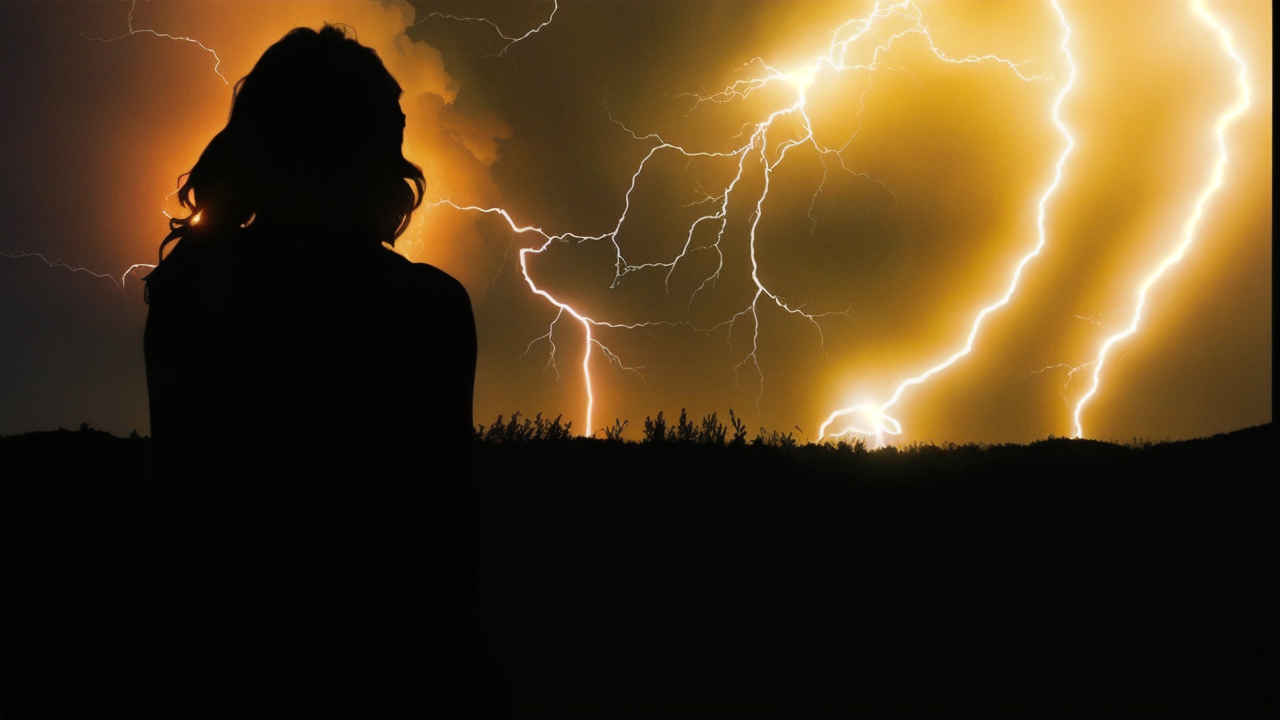} &
\includegraphics[width=\figurewidthnine\linewidth]{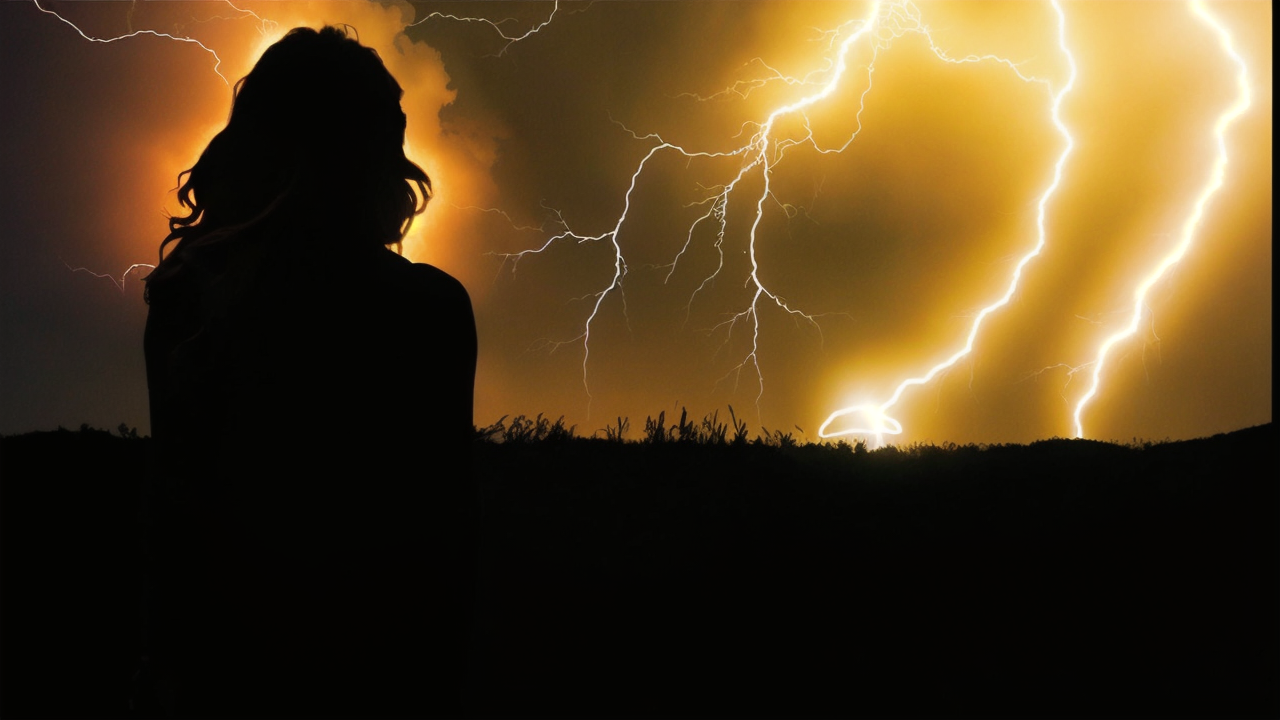} &
\includegraphics[width=\figurewidthnine\linewidth]{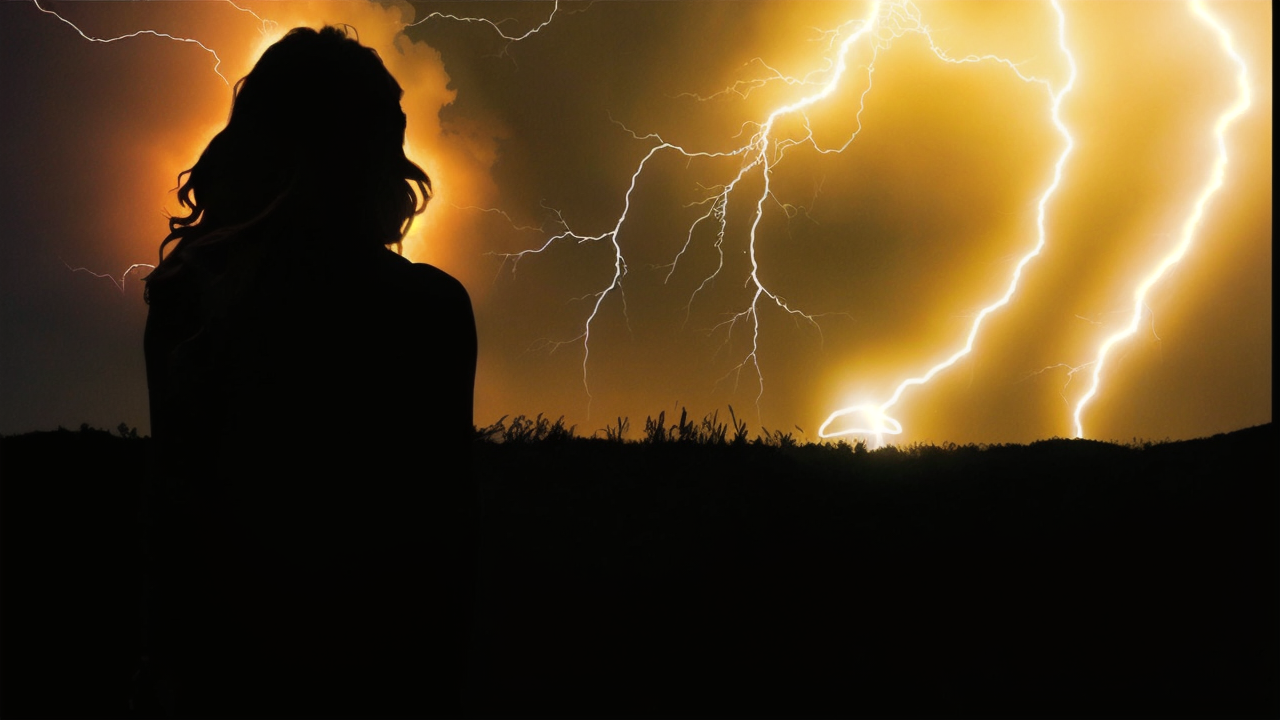} &
\includegraphics[width=\figurewidthnine\linewidth]{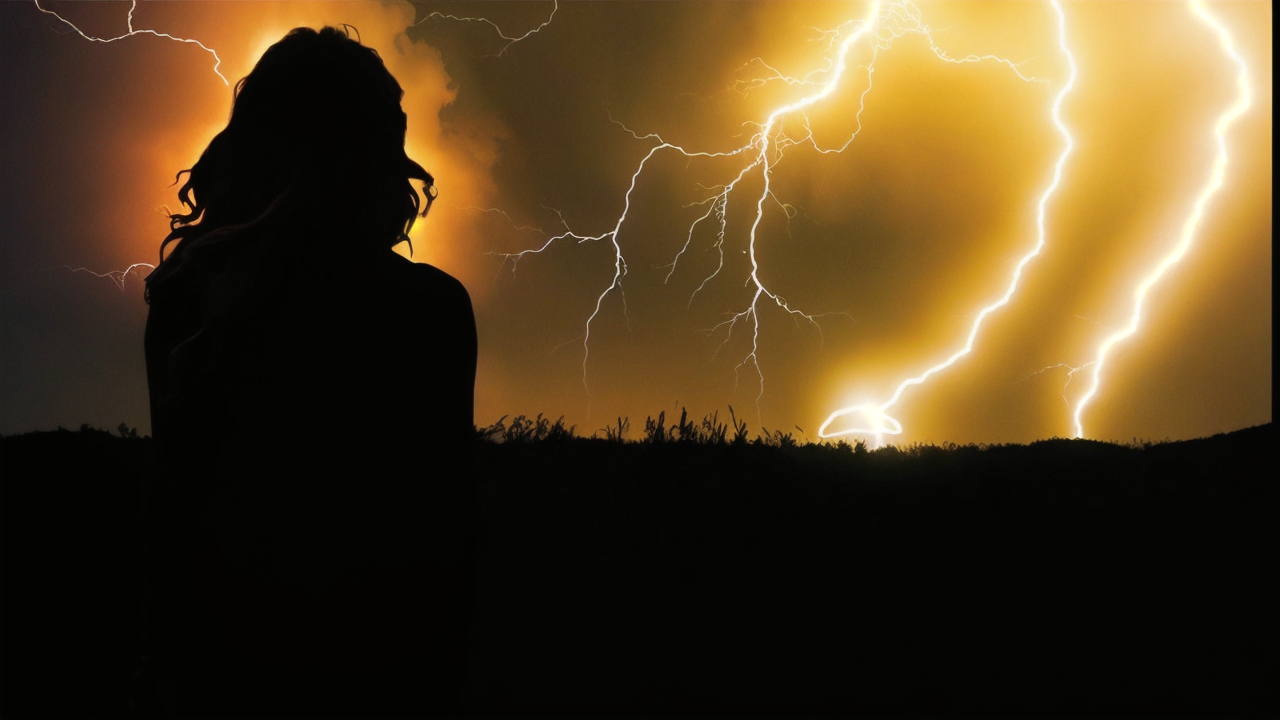} &
\includegraphics[width=\figurewidthnine\linewidth]{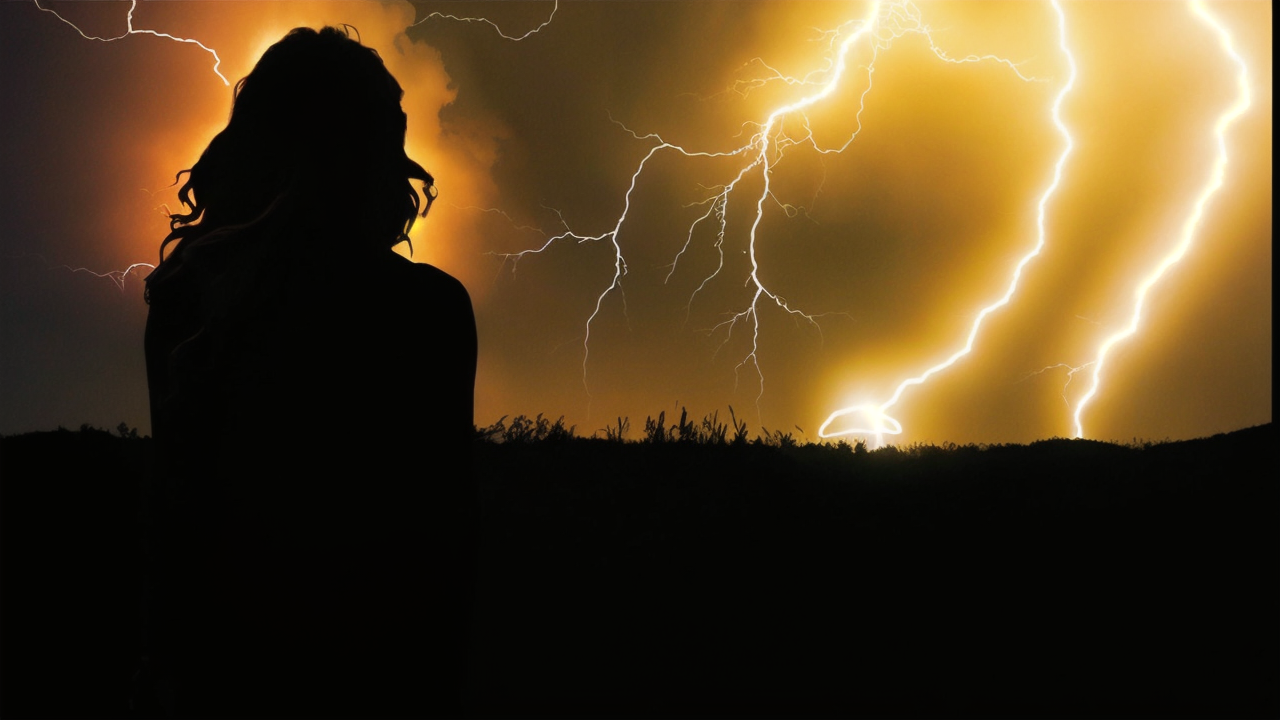} &
\includegraphics[width=\figurewidthnine\linewidth]{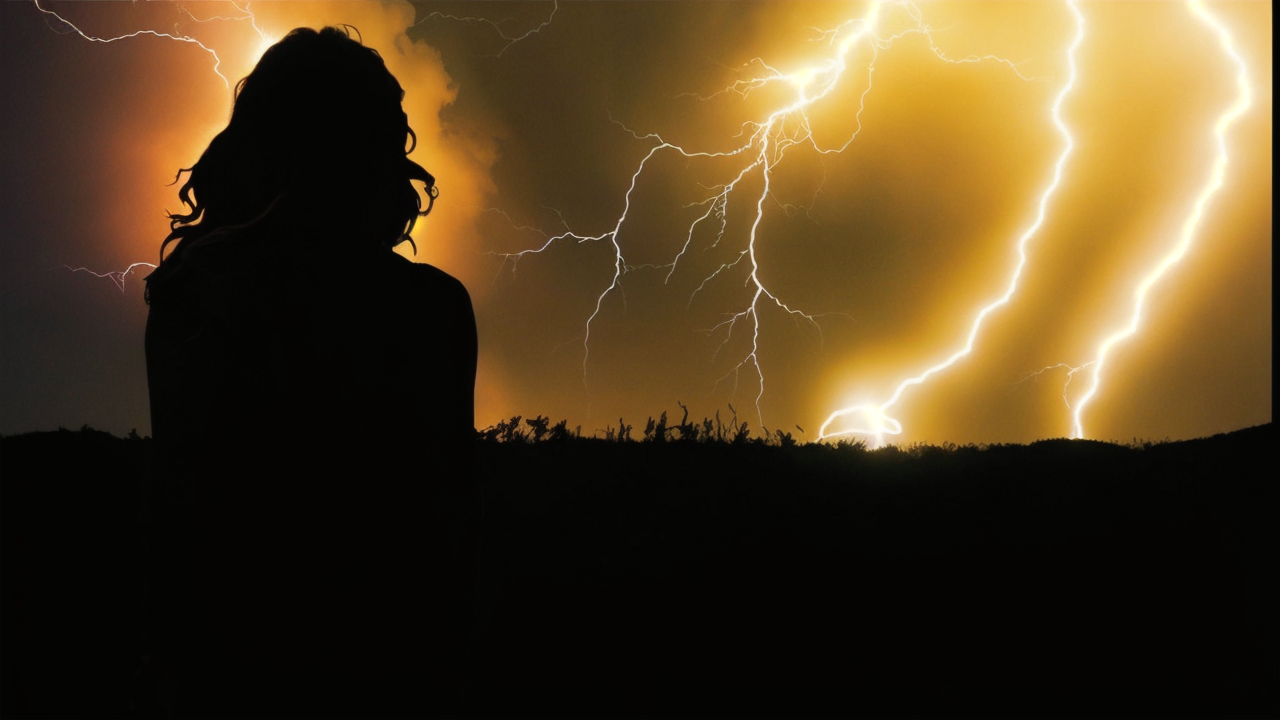} &
\includegraphics[width=\figurewidthnine\linewidth]{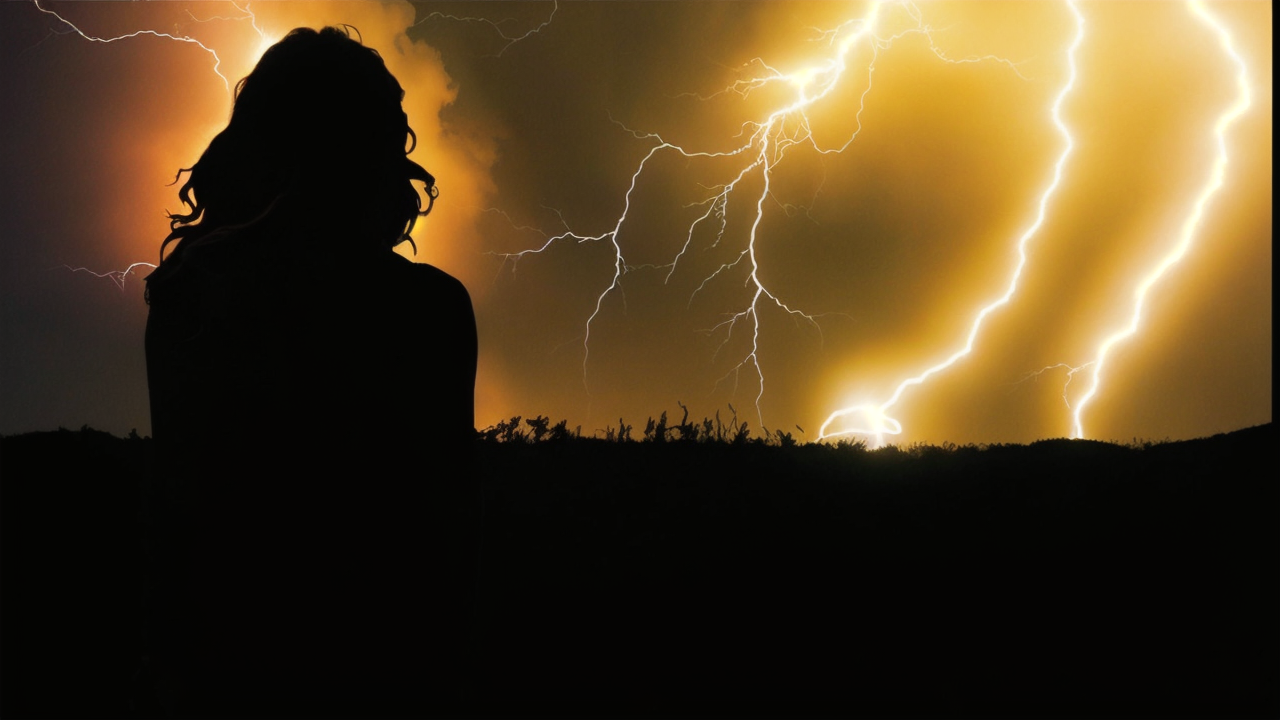} &
\includegraphics[width=\figurewidthnine\linewidth]{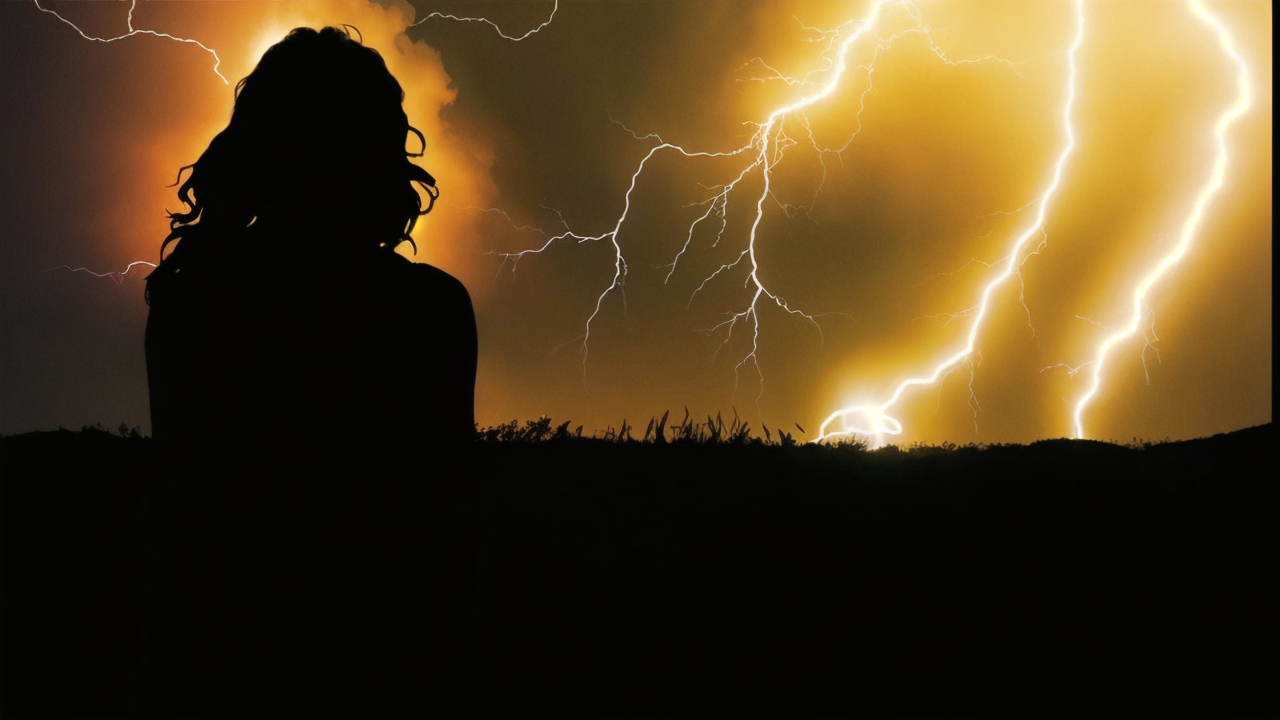} &
\includegraphics[width=\figurewidthnine\linewidth]{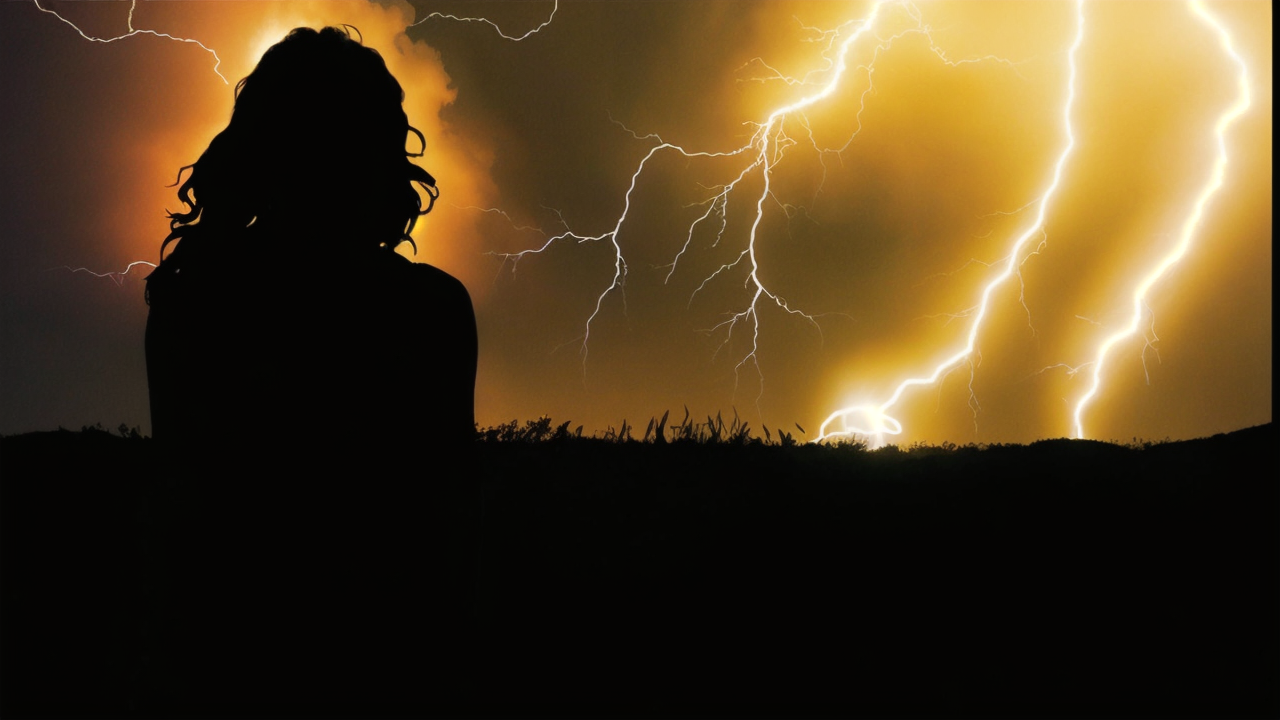} &
... \\
\multicolumn{11}{c}{(b) Diffusion-rendered video} \\
&
\multicolumn{3}{c}{\includegraphics[width=\figurewidththree\linewidth]{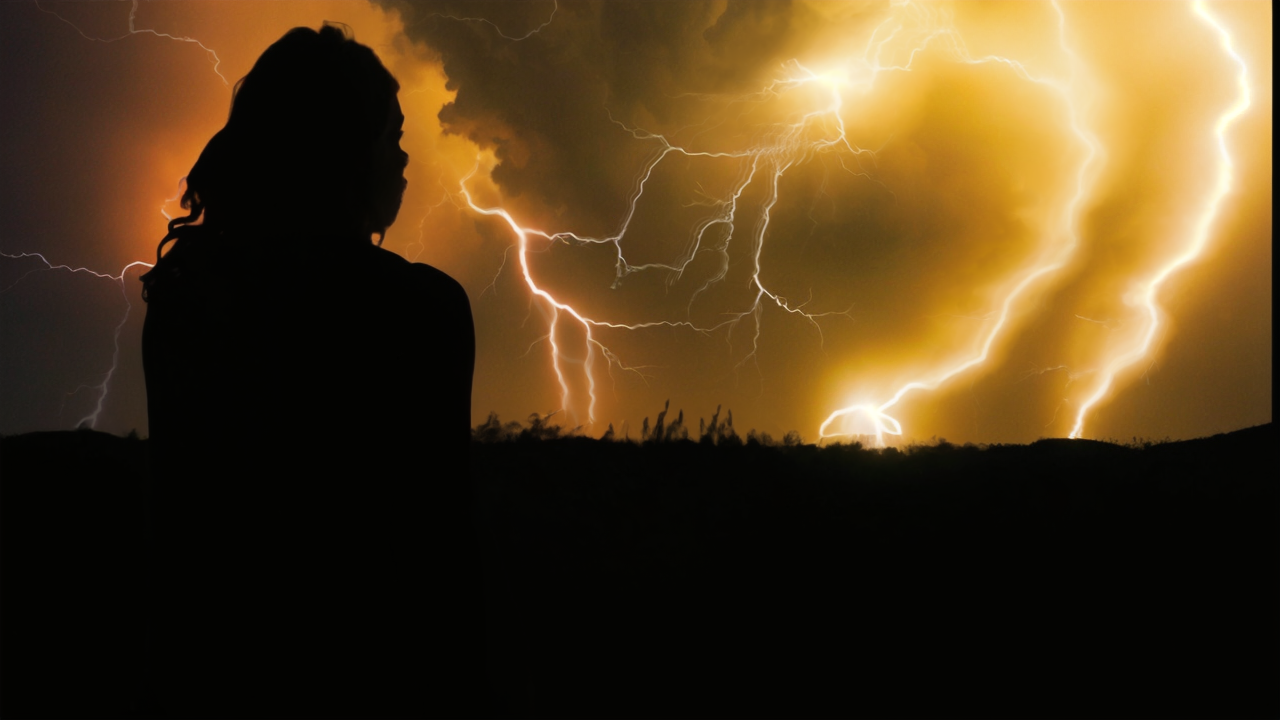}} &
\multicolumn{3}{c}{\includegraphics[width=\figurewidththree\linewidth]{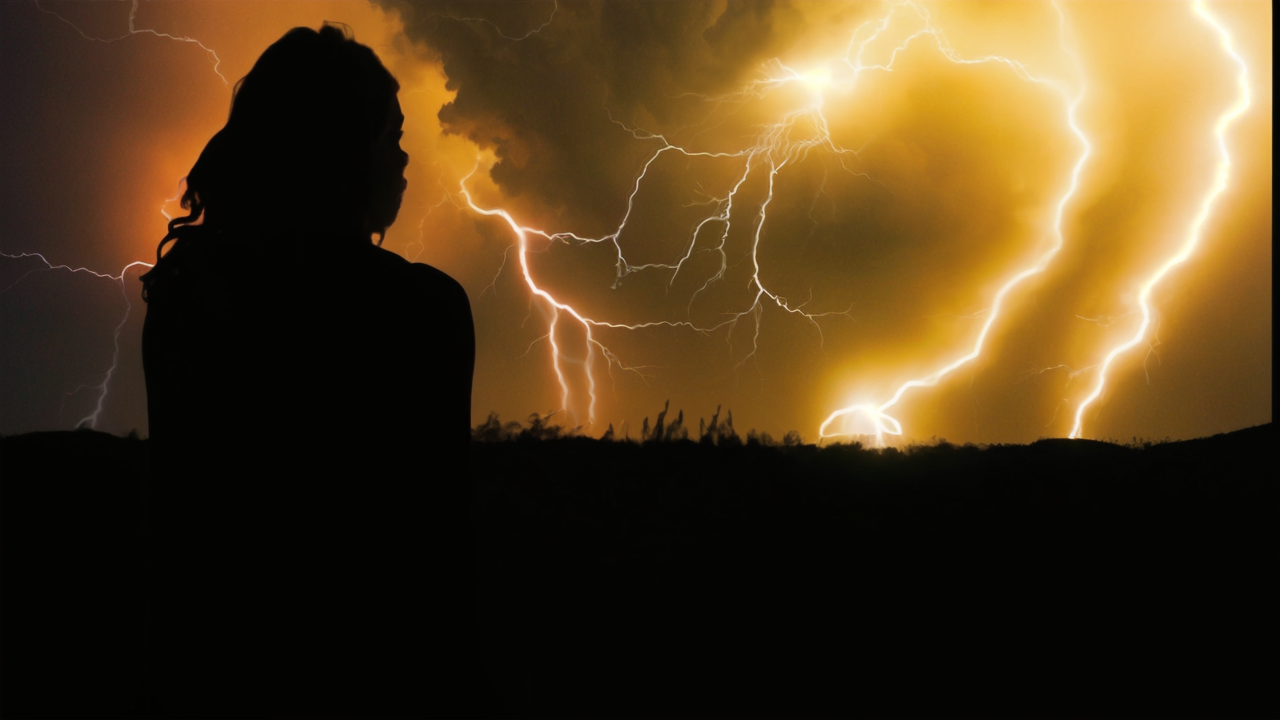}} &
\multicolumn{3}{c}{\includegraphics[width=\figurewidththree\linewidth]{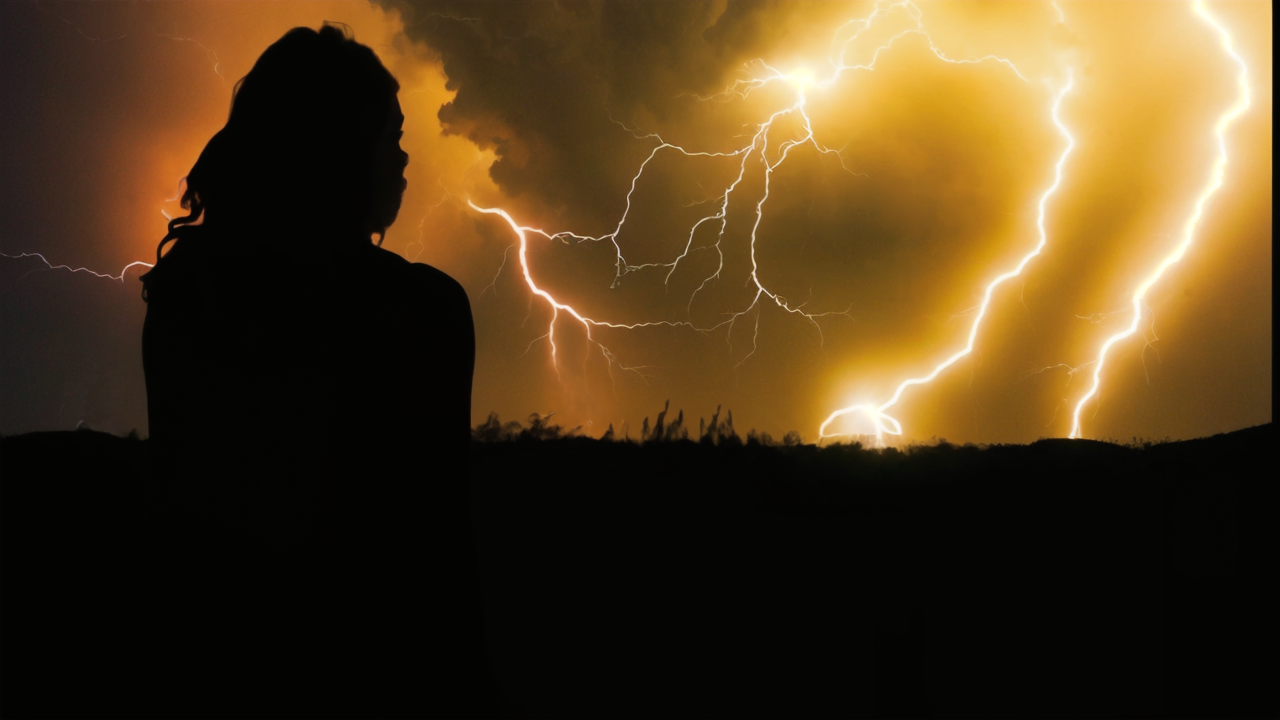}} &
\\
&
\multicolumn{3}{c}{(c) Fast mode} &
\multicolumn{3}{c}{(d) Balanced mode} &
\multicolumn{3}{c}{(e) Accurate mode} &
\\
\end{tabular}
\caption{Comparison of the three inference modes for blending. Zoom in to see details.}
\label{figure:blending_mode}
\end{figure*}

\newcommand{\figurewidthtwo}{0.47}
\begin{figure}[]
\centering
\tabcolsep=3pt
\begin{tabular}{cc}
\includegraphics[width=\figurewidthtwo\linewidth]{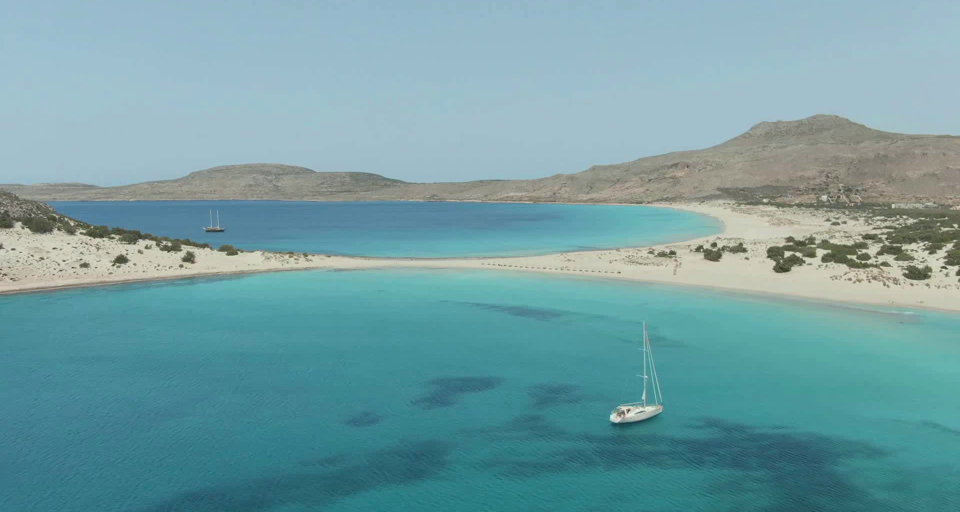} &
\includegraphics[width=\figurewidthtwo\linewidth]{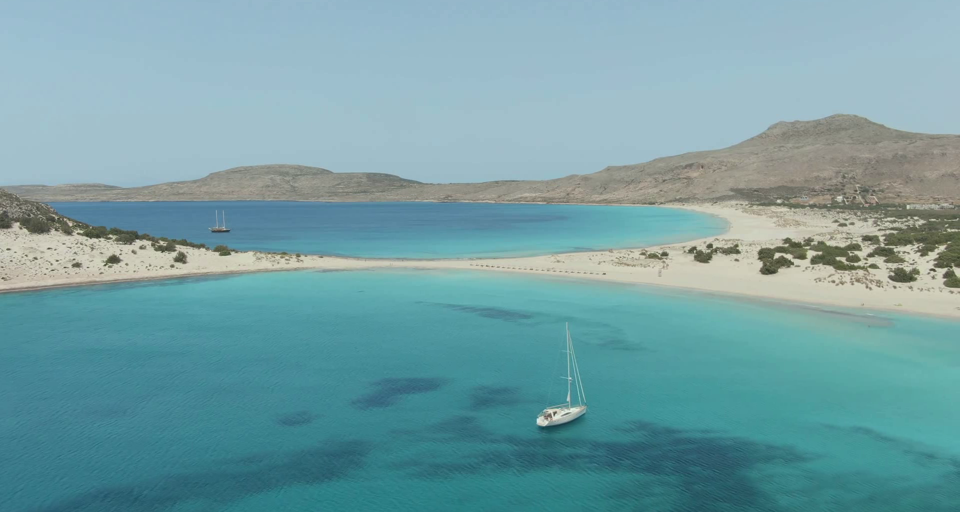} \\
\multicolumn{2}{c}{(a) Original video} \\
\includegraphics[width=\figurewidthtwo\linewidth]{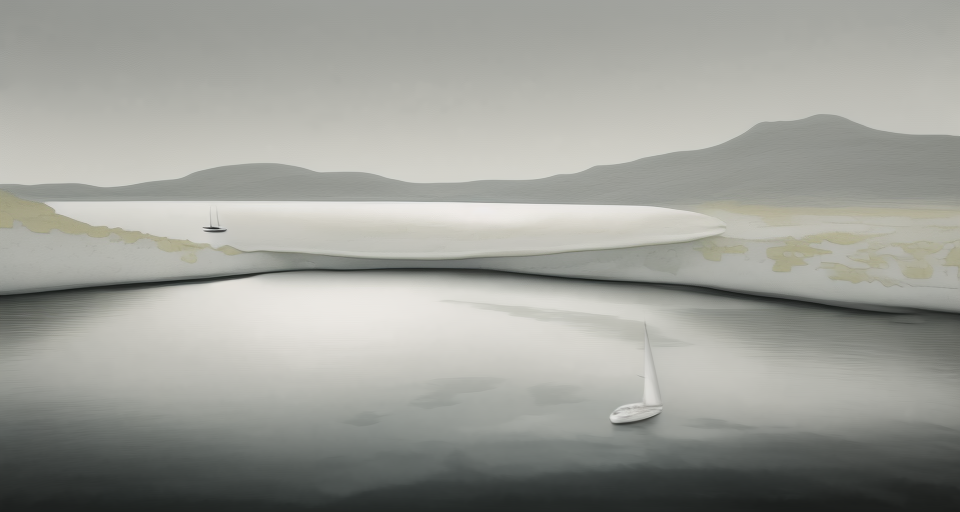} &
\includegraphics[width=\figurewidthtwo\linewidth]{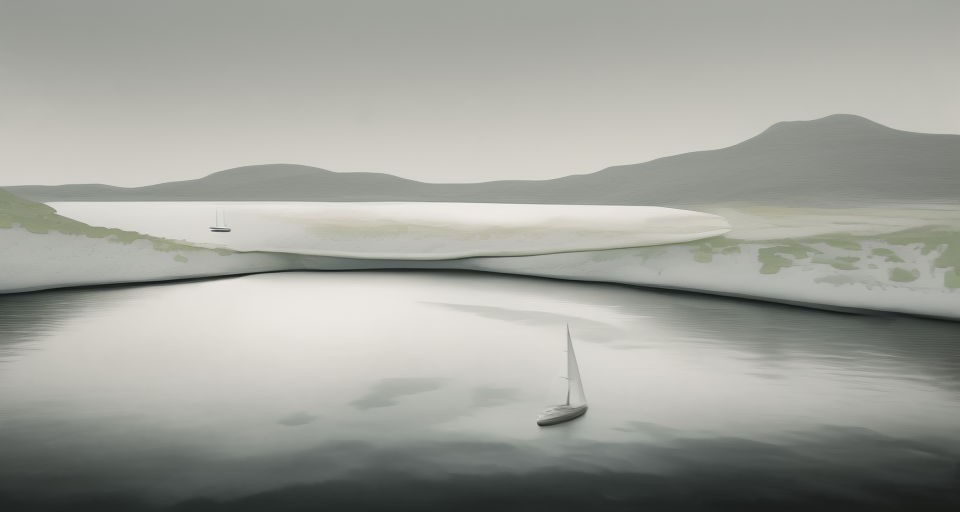} \\
\multicolumn{2}{c}{(b) The size of sliding window is 30} \\
\includegraphics[width=\figurewidthtwo\linewidth]{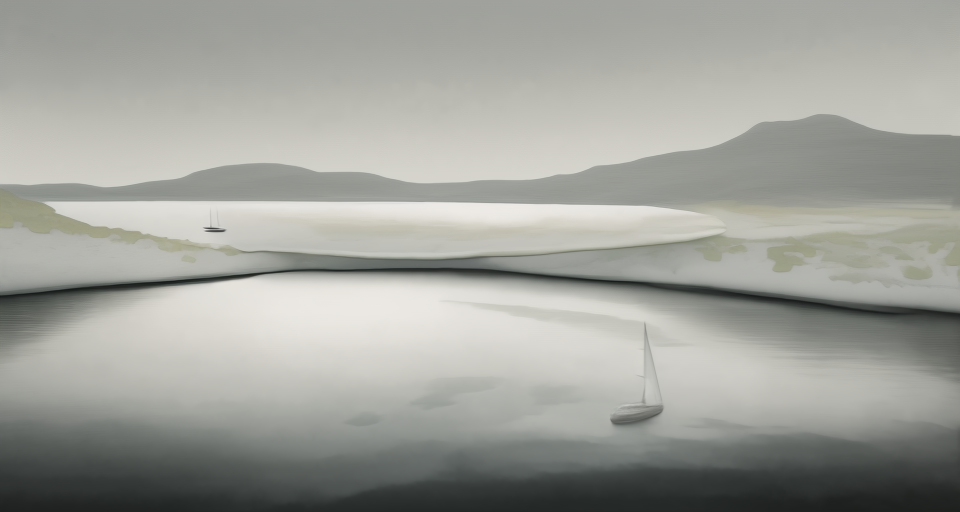} &
\includegraphics[width=\figurewidthtwo\linewidth]{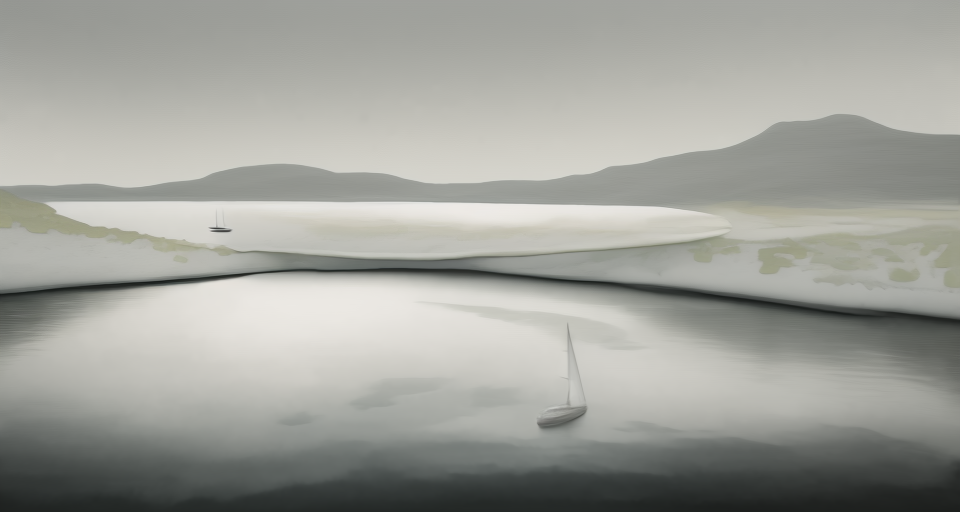} \\
\multicolumn{2}{c}{(b) The sliding window covers the whole video} \\
\end{tabular}
\caption{Effect of different sizes of the sliding window.}
\label{figure:window_size}
\end{figure}

\begin{figure}[]
\centering
\tabcolsep=3pt
\begin{tabular}{cc}
\includegraphics[width=\figurewidthtwo\linewidth]{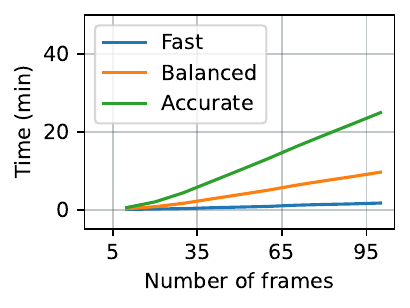} &
\includegraphics[width=\figurewidthtwo\linewidth]{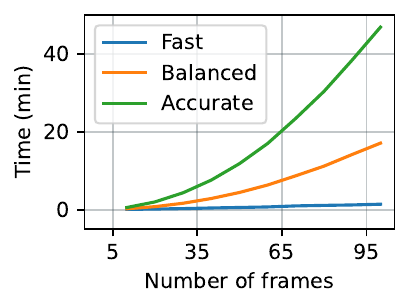} \\
(a) The size of sliding &
(b) The sliding window\\
window is 30 &
covers the whole video\\
\end{tabular}
\caption{Inference time of different inference modes for blending.}
\label{figure:efficiency}
\end{figure}

We compared the performance of different inference modes in \ref{subsection:blending}. Figure \ref{figure:blending_mode} displays the results of three inference modes. The positions of lightning in the video are random and inconsistent. In the balanced mode (Figure \ref{figure:blending_mode}d), there is a slight ghosting effect where the lightning from different frames is blended together. The ghosting issue is more pronounced in the fast mode (Figure \ref{figure:blending_mode}c). However, in the accurate mode (Figure \ref{figure:blending_mode}e), loss function \ref{equation:loss_function_3} guides the optimization algorithm to remove unnecessary details and merge lightning from multiple frames into a clear frame.

We also conducted experiments to investigate the impact of different sliding window sizes. Figure \ref{figure:window_size} shows the first and last frames of a 125-frame video. When the sliding window size is set to 30 (Figure \ref{figure:window_size}b), the color of the boat in the scene is different because the two frames are far apart. As the sliding window covers the entire video (Figure \ref{figure:window_size}c), the color of the small boat becomes consistent. Larger sliding window sizes can improve the long-term consistency of videos but require more computation time.

Figure \ref{figure:efficiency} records the computation time for three inference modes. The GPU used is an NVIDIA 4090, and the video resolution is $512\times 768$. For the balanced and accurate modes, when the sliding window size is 30, the computation time almost linearly increases with the number of frames, and when the sliding window covers the entire video, the computation time increases quadratically with the number of frames. For the fast mode, due to the linear logarithmic time complexity, the speed is very fast and is almost unaffected by the sliding window size. We also tested on a GPU with very low computing performance. We conducted tests on an NVIDIA 3060 laptop, a GPU that doesn't even have enough VRAM to run baseline methods like Text2Video-Zero \cite{khachatryan2023text2video} and CoDeF \cite{ouyang2023codef}. However, FastBlend only takes 8 minutes to process 200 frames in the fast mode.

\subsection{Ablation Study}

\subsubsection{Tracking}

\begin{figure*}[]
\centering
\tabcolsep=3pt
\begin{tabular}{cccc}
\includegraphics[width=\figurewidth\linewidth]{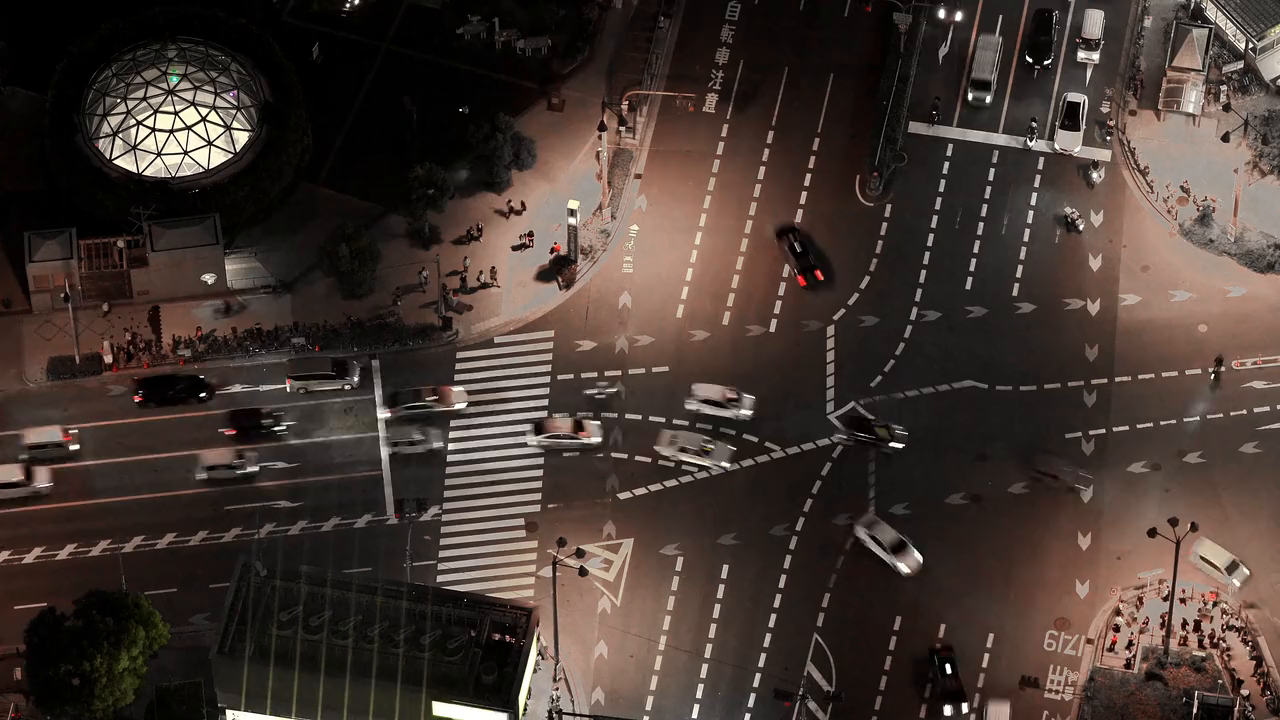} &
\includegraphics[width=\figurewidth\linewidth]{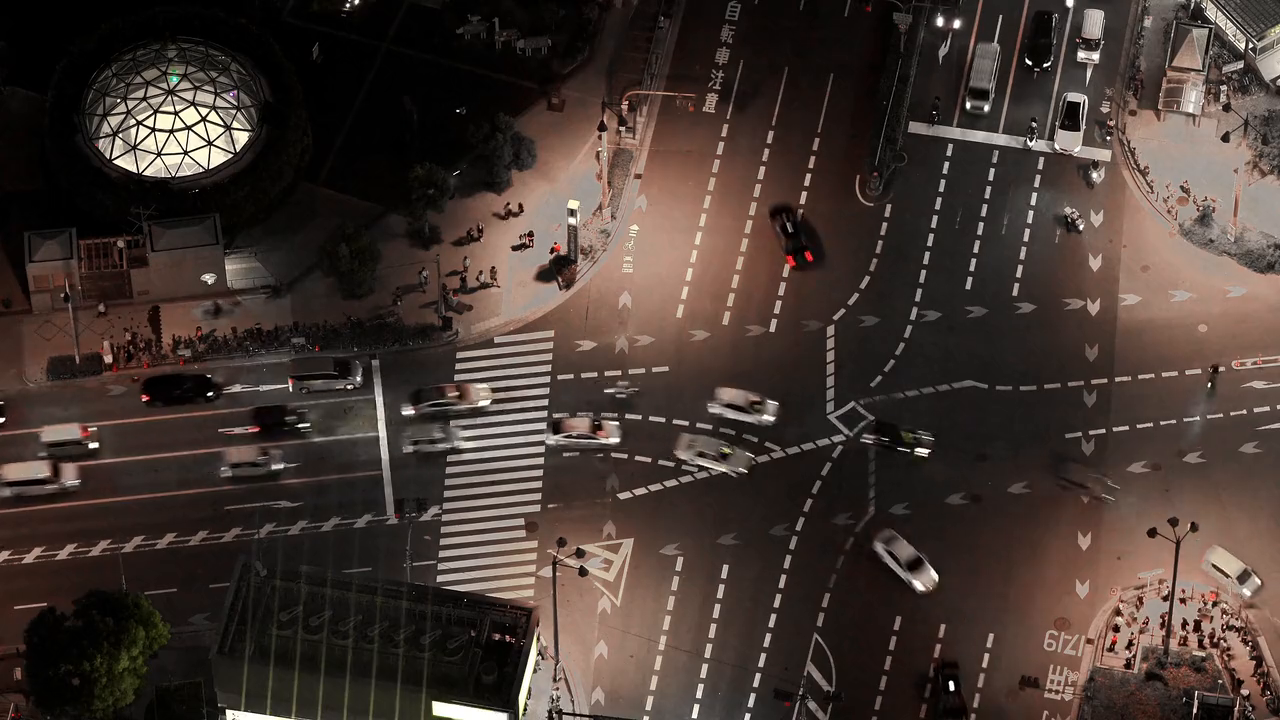} &
\includegraphics[width=\figurewidth\linewidth]{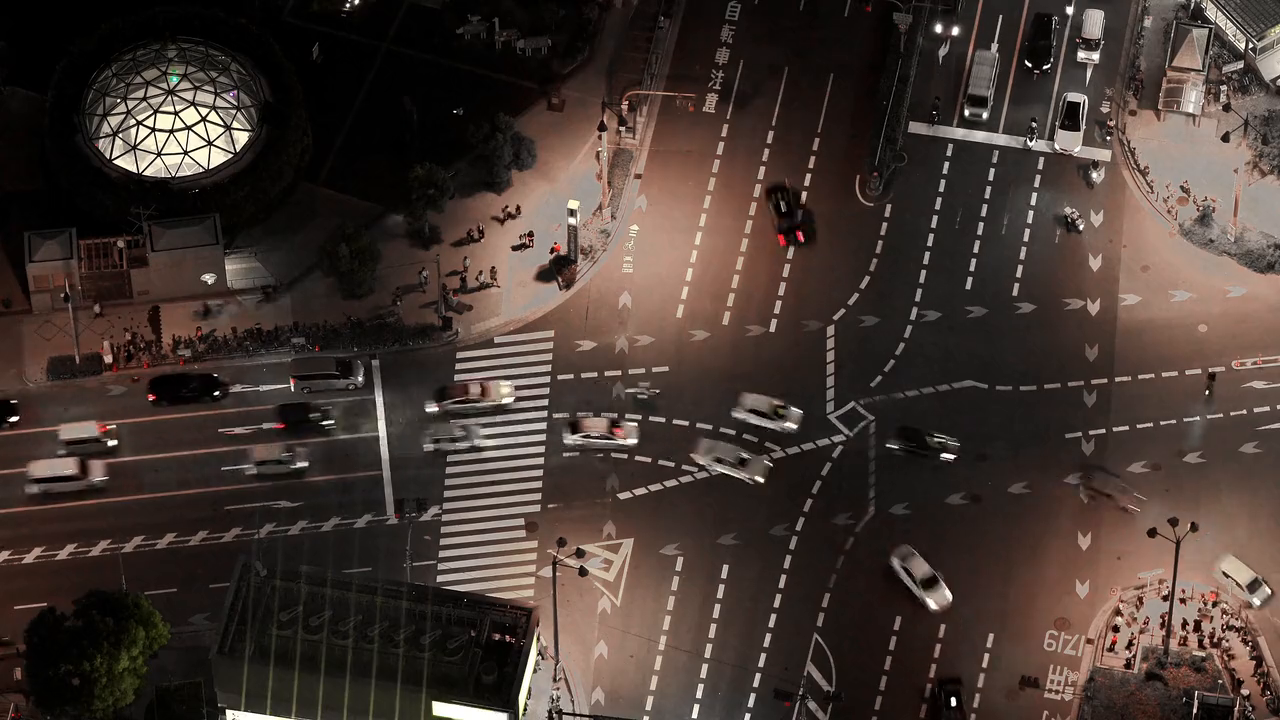} &
\includegraphics[width=\figurewidth\linewidth]{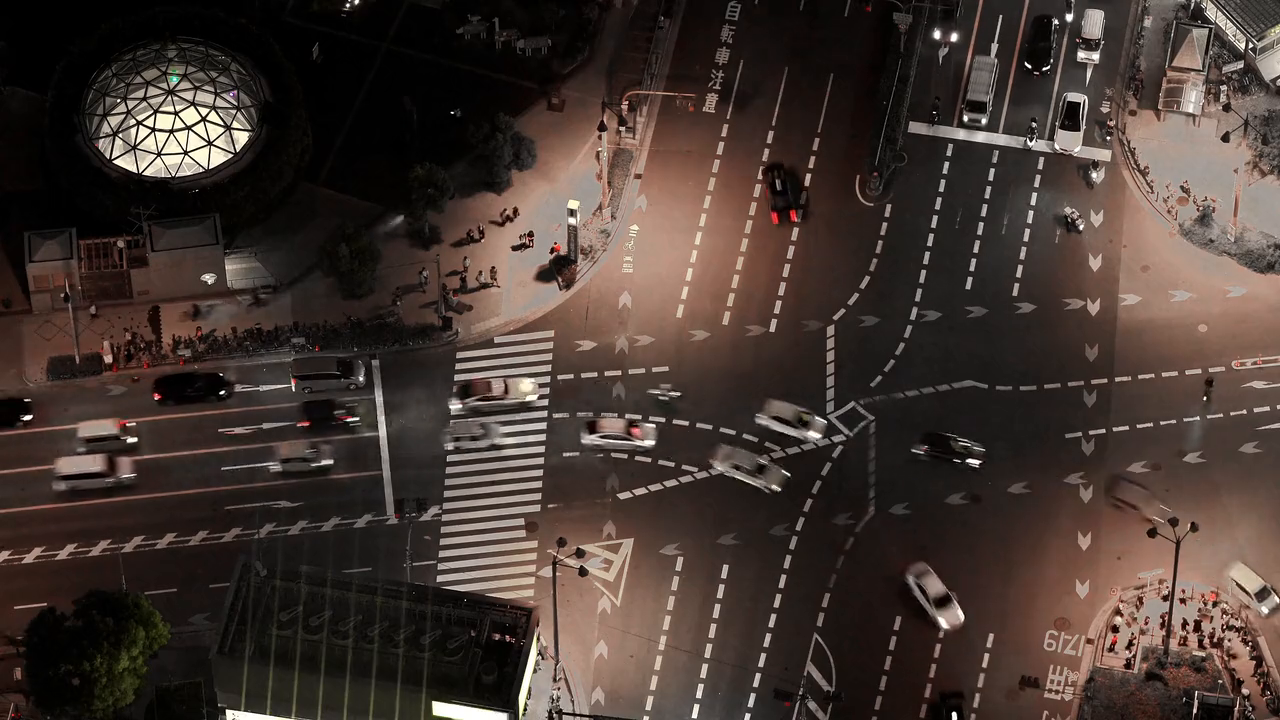} \\
\multicolumn{4}{c}{(a) Original video} \\
\includegraphics[width=\figurewidth\linewidth]{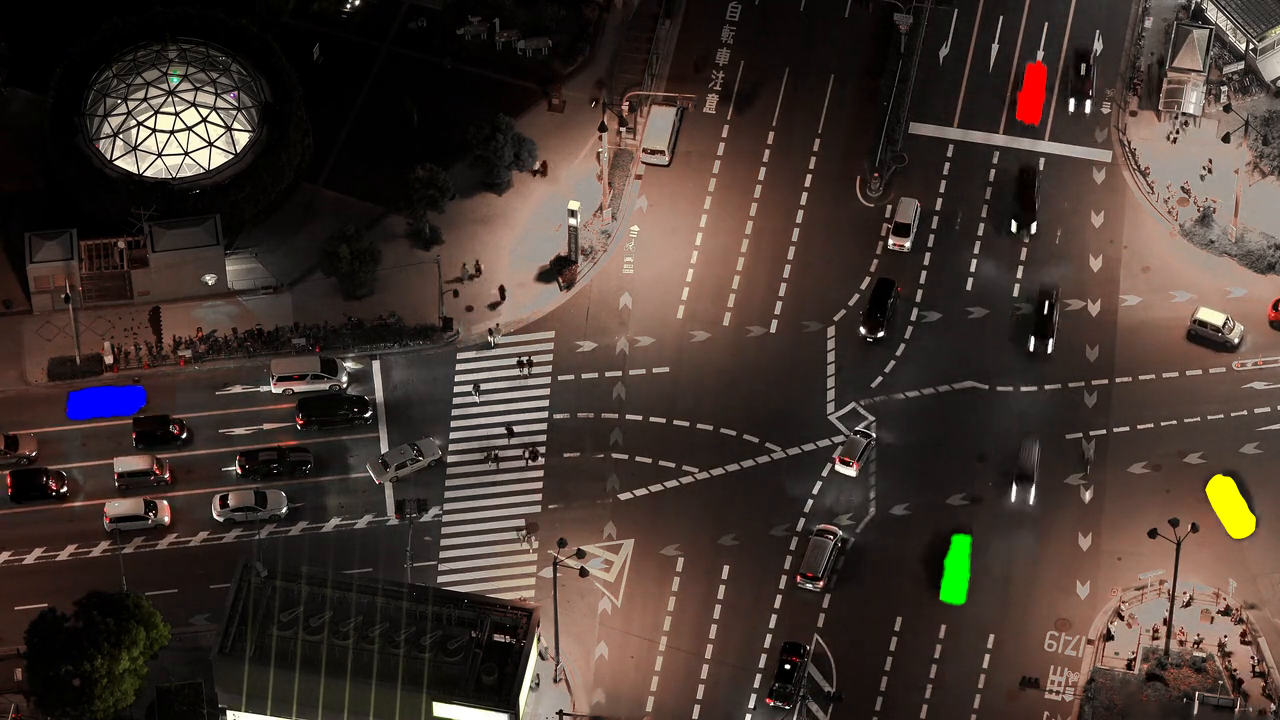} &
\includegraphics[width=\figurewidth\linewidth]{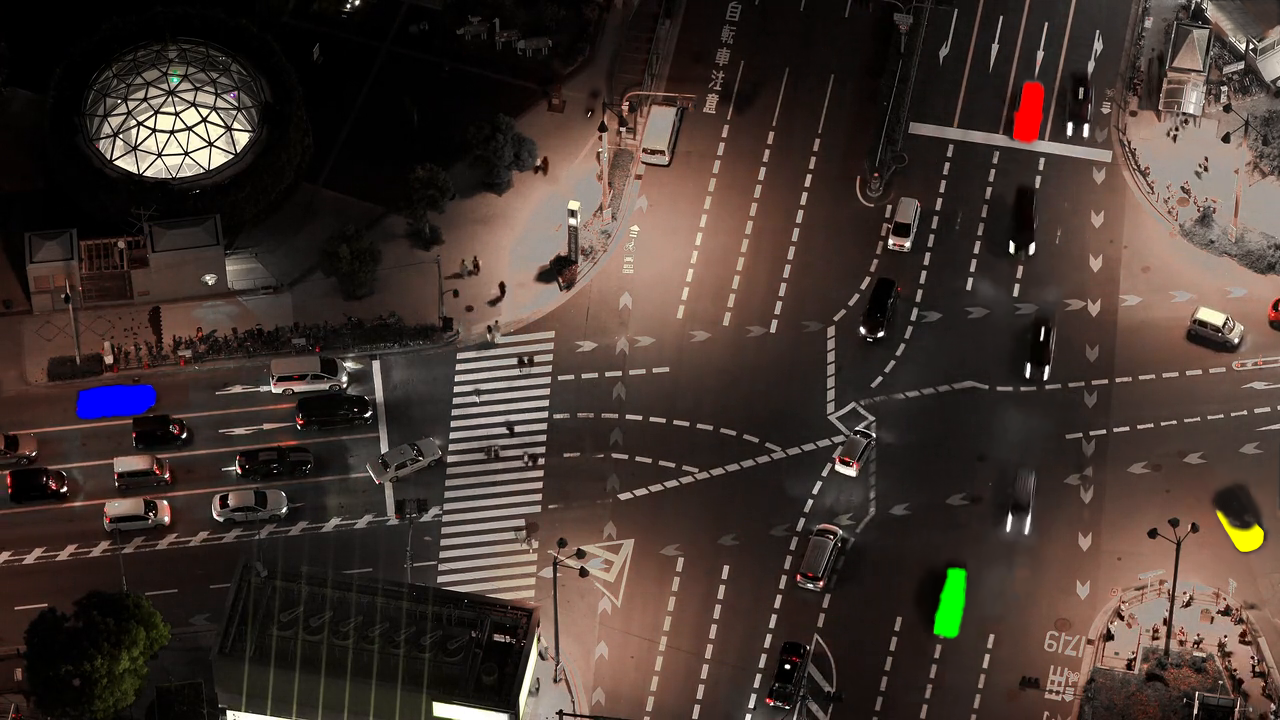} &
\includegraphics[width=\figurewidth\linewidth]{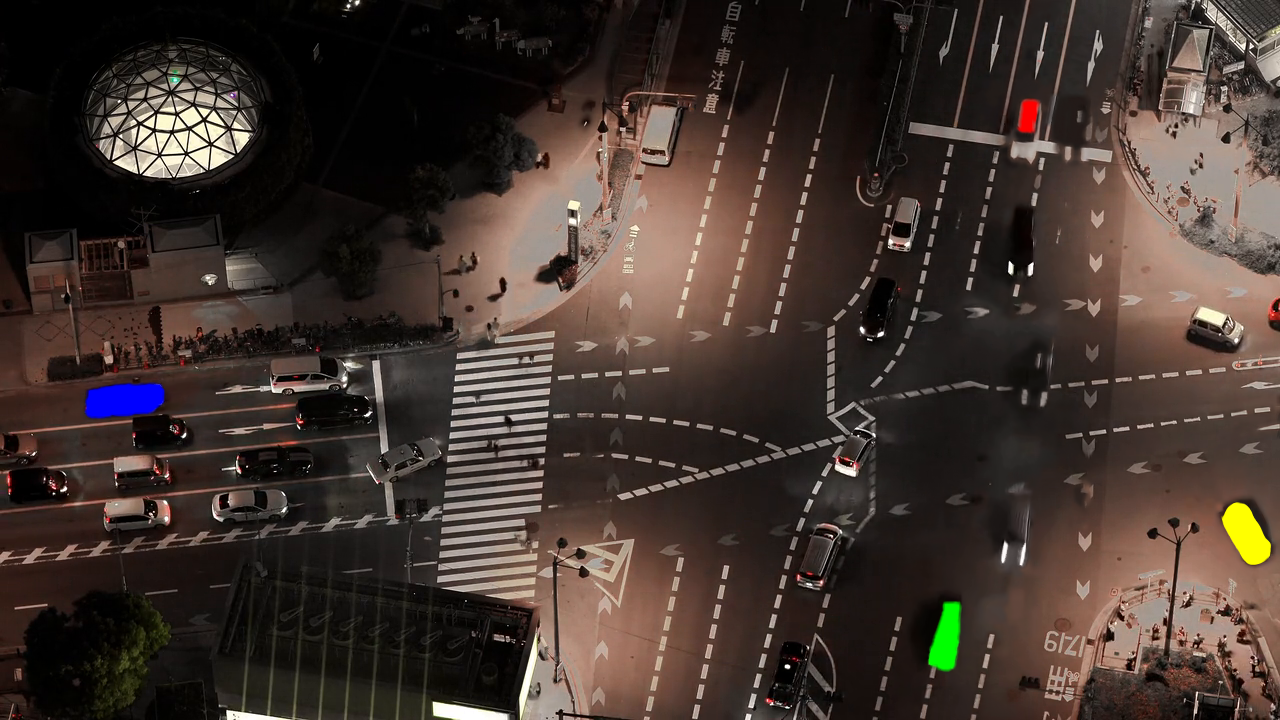} &
\includegraphics[width=\figurewidth\linewidth]{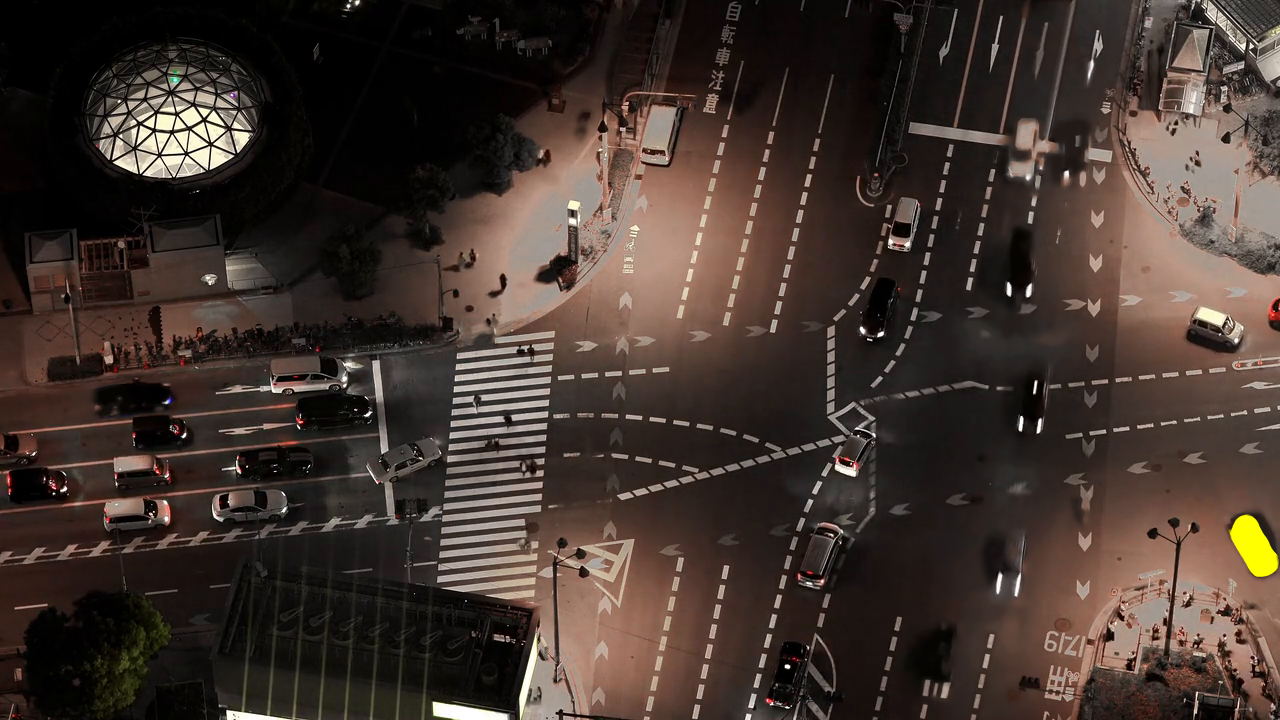} \\
\multicolumn{4}{c}{(b) Disabling the object tracking mechanism} \\
\includegraphics[width=\figurewidth\linewidth]{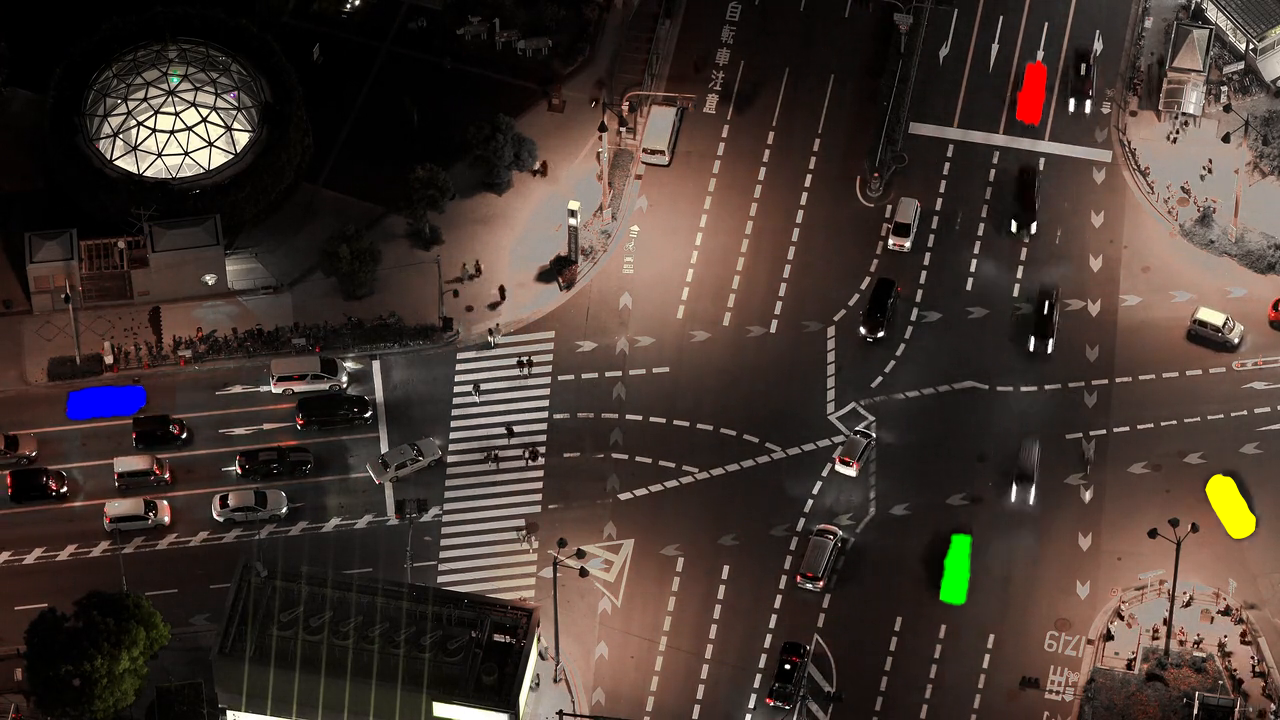} &
\includegraphics[width=\figurewidth\linewidth]{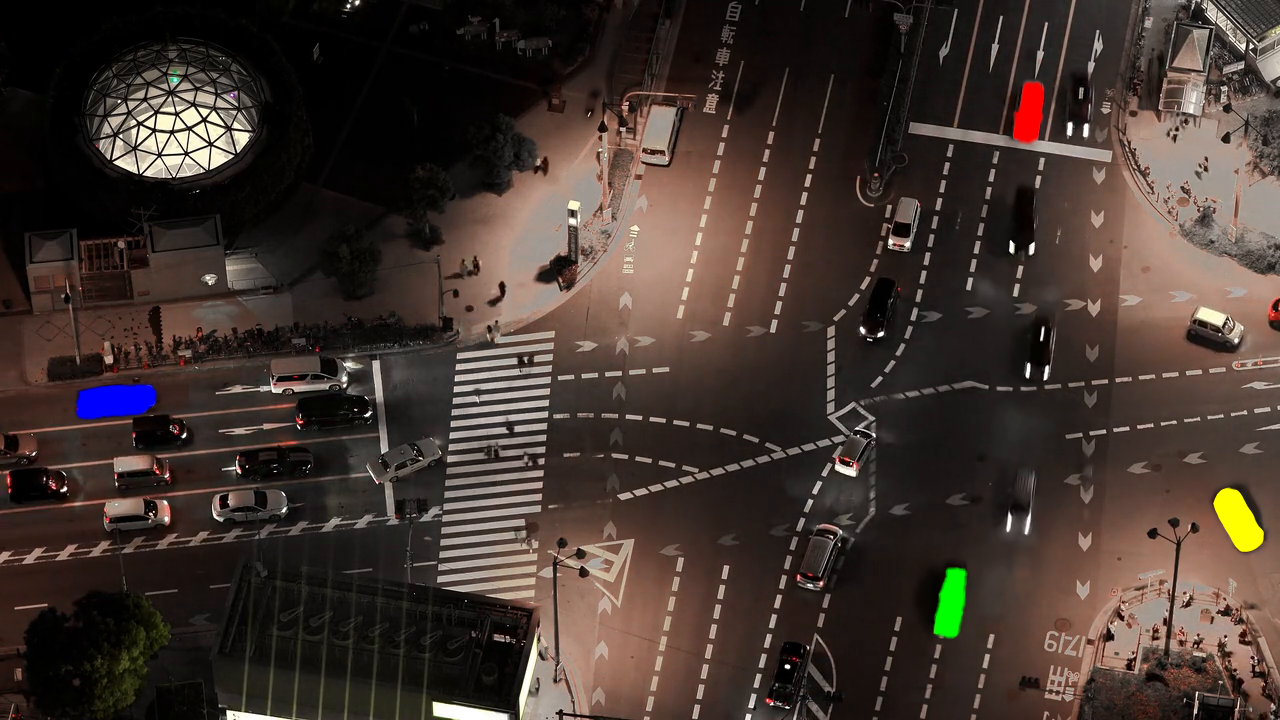} &
\includegraphics[width=\figurewidth\linewidth]{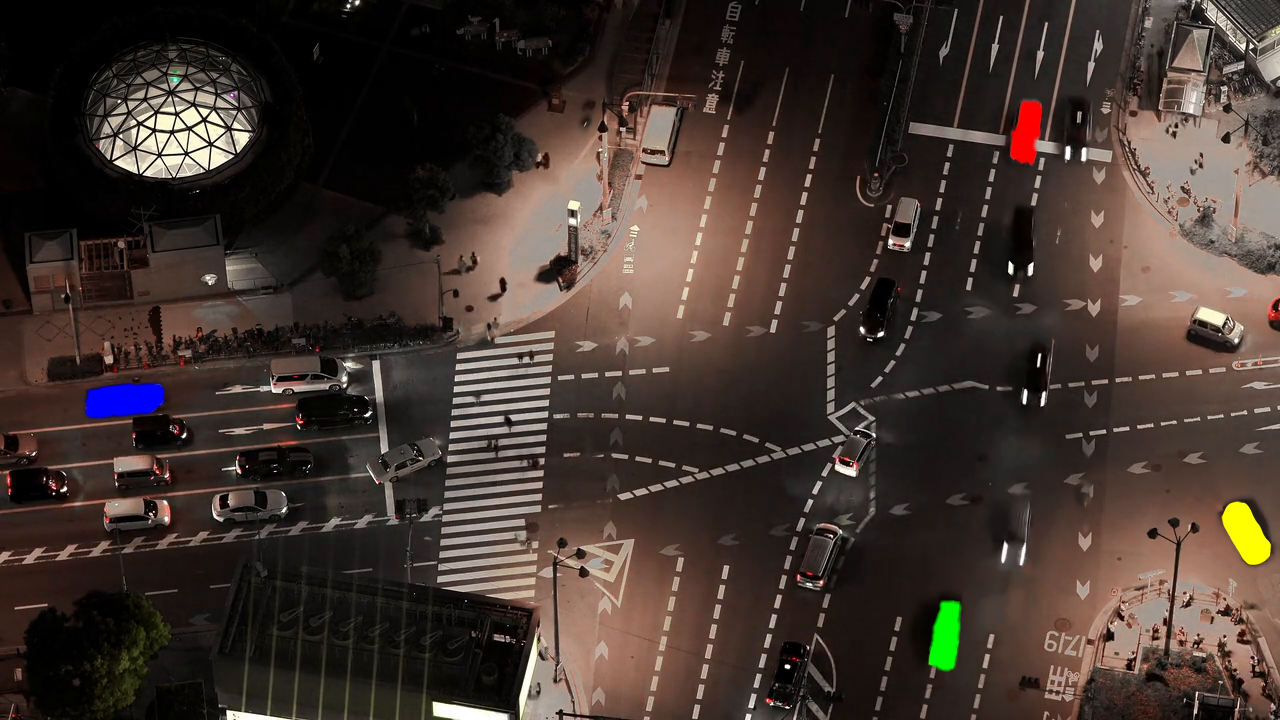} &
\includegraphics[width=\figurewidth\linewidth]{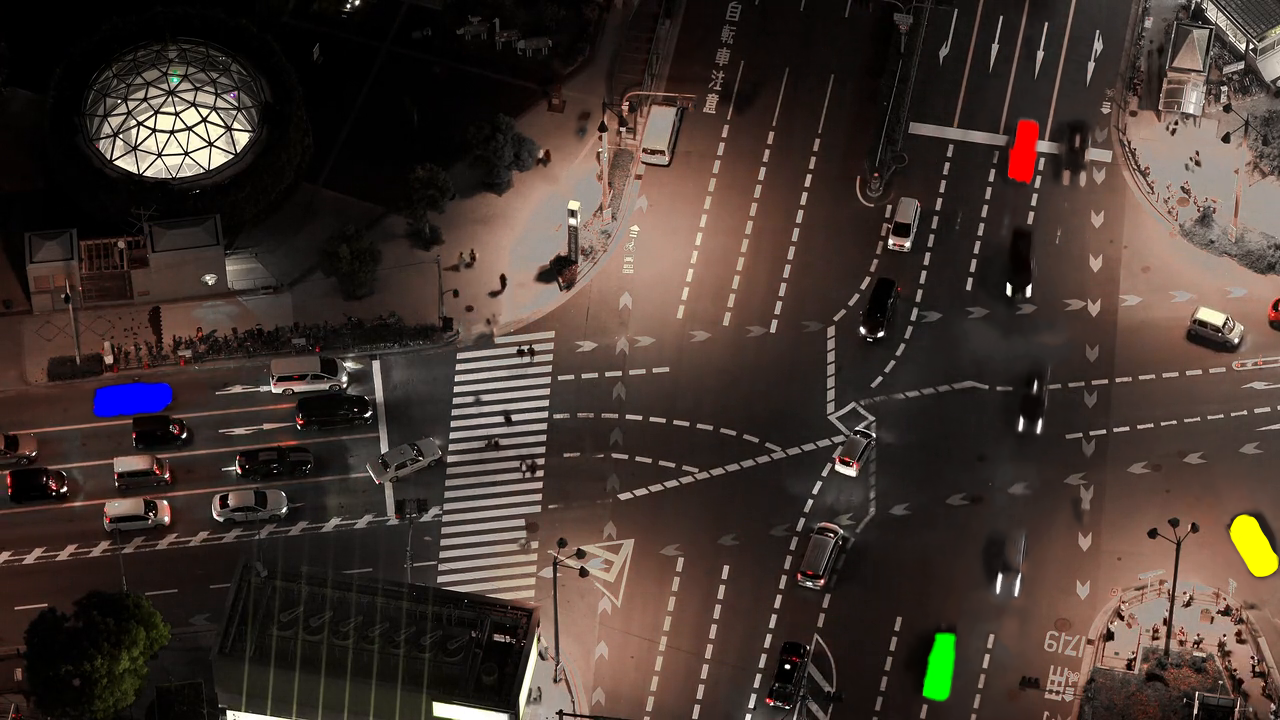} \\
\multicolumn{4}{c}{(c) Enabling the object tracking mechanism} \\
\end{tabular}
\caption{Effect of the object tracking mechanism.}
\label{figure:tracking}
\end{figure*}

We evaluate the object tracking mechanism in \ref{subsubsection:tracking}. Figure \ref{figure:tracking} shows a video of a congested intersection. In the original video (Figure \ref{subsubsection:tracking}a), there are many vehicles. In the first frame, we select four moving cars and color them differently, and then we use the first frame as a keyframe to render this video. When the object tracking mechanism is disabled (Figure \ref{subsubsection:tracking}b), the colors of the four cars quickly fade away. However, when the object tracking mechanism is enabled (Figure \ref{subsubsection:tracking}c), the colors on the four cars remain stable. This tracking mechanism can enhance the stability of rendering videos. However, this mechanism is only designed for NNF estimation and is not a general-purpose tracking algorithm. We do not recommend using this tracking mechanism for general tracking tasks, because it is not widely evaluated on general tracking tasks and is less robust compared to some prior general approaches \cite{wojke2017simple, huang2020globaltrack}.

\subsubsection{Alignment}

\begin{figure}[]
\centering
\tabcolsep=3pt
\begin{tabular}{cc}
\includegraphics[width=\figurewidthtwo\linewidth]{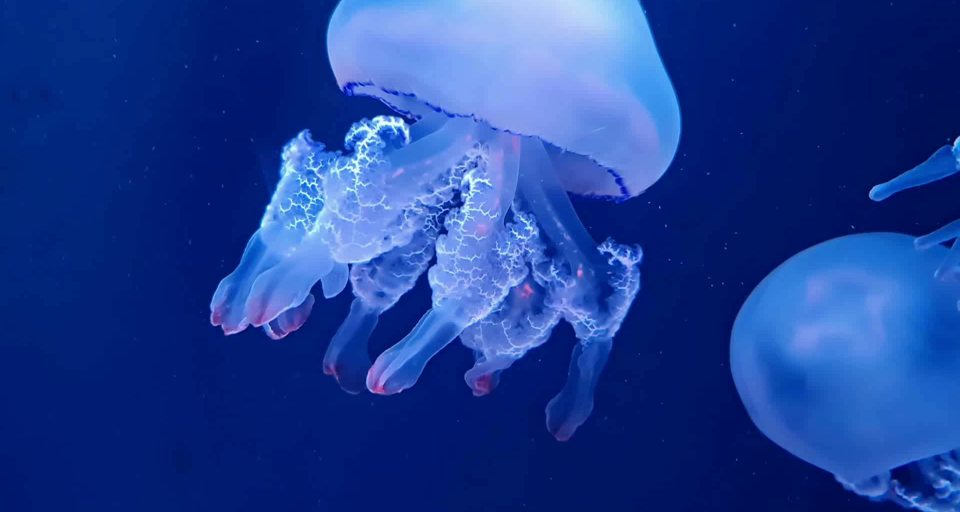} &
\includegraphics[width=\figurewidthtwo\linewidth]{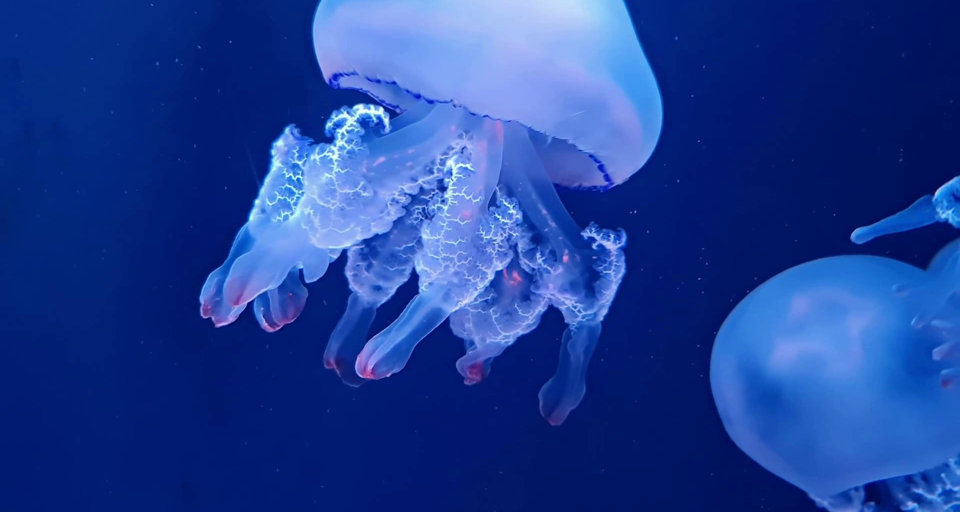} \\
\multicolumn{2}{c}{(a) Keyframes} \\
\includegraphics[width=\figurewidthtwo\linewidth]{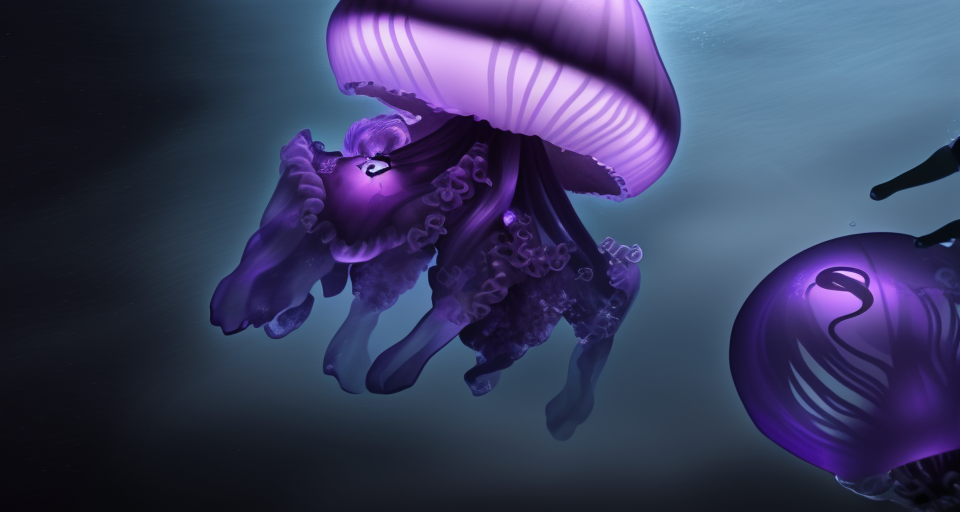} &
\includegraphics[width=\figurewidthtwo\linewidth]{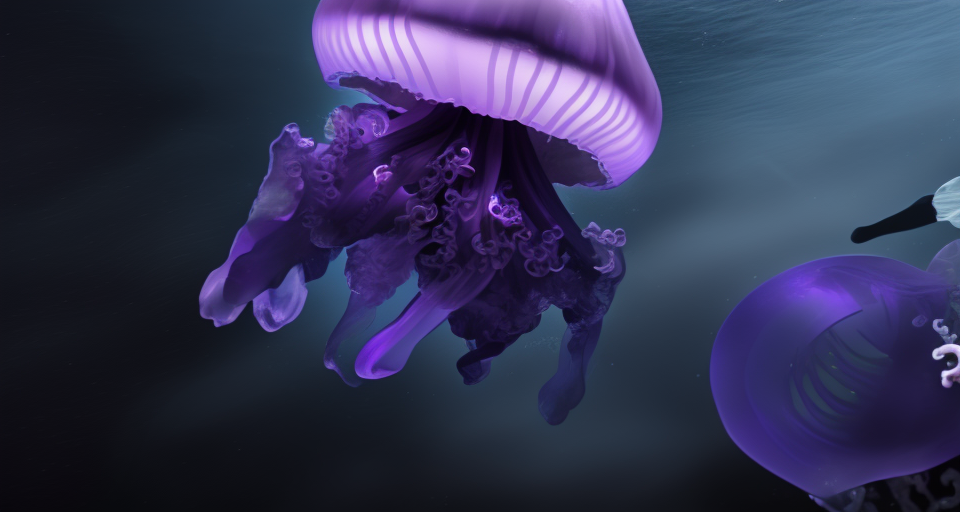} \\
\multicolumn{2}{c}{(b) Diffusion-rendered keyframes} \\

\begin{tikzpicture}
    \node[anchor=south west,inner sep=0] (image) at (0,0) {\includegraphics[width=\figurewidthtwo\linewidth]{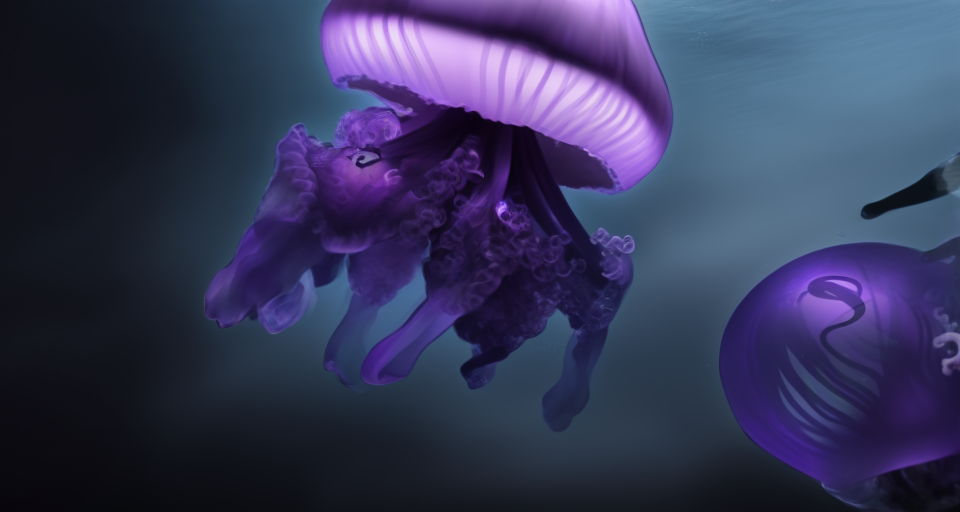}};
    \begin{scope}[x={(image.south east)},y={(image.north west)}]
        \draw[red, ultra thick, rounded corners] (0.8,0.2) rectangle (0.95,0.55);
    \end{scope}
\end{tikzpicture} &

\begin{tikzpicture}
    \node[anchor=south west,inner sep=0] (image) at (0,0) {\includegraphics[width=\figurewidthtwo\linewidth]{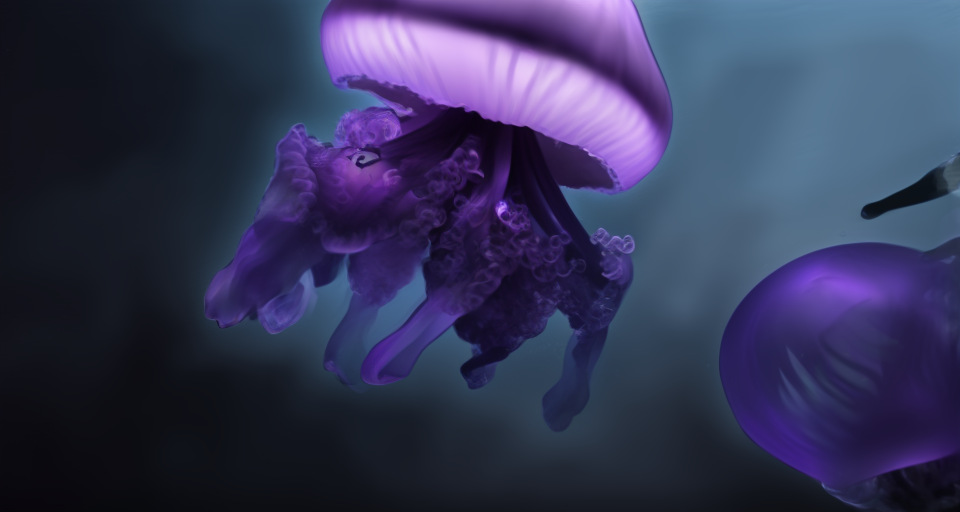}};
    \begin{scope}[x={(image.south east)},y={(image.north west)}]
        \draw[red, ultra thick, rounded corners] (0.8,0.2) rectangle (0.95,0.55);
    \end{scope}
\end{tikzpicture} \\
(c) Interpolated frame &
(d) Interpolated frame\\
without alignment &
with alignment\\
\end{tabular}
\caption{Effect of the alignment mechanism.}
\label{figure:alignment}
\end{figure}

Figure \ref{figure:alignment} illustrates the effect of the alignment mechanism in \ref{subsubsection:alignment}. After rendering the jellyfish in two keyframes (Figure \ref{figure:alignment}a) into purple (Figure \ref{figure:alignment}b), inconsistent content appears in the bottom right corner of the video. When we generate the intermediate frame in interpolation mode without the alignment mechanism (Figure \ref{figure:alignment}c), the inconsistent content transparently overlays. When we generate the intermediate frames with the alignment mechanism, the inconsistent content is eliminated. The alignment mechanism effectively integrates the content from two inconsistent keyframes.

\section{Conclusion}

In this paper, we propose a model-free video processing toolkit called FastBlend. FastBlend can be combined with diffusion models to create powerful video processing pipelines. FastBlend can effectively eliminate flicker in videos, interpolate keyframe sequences, and even process complete videos based on a single image. Extensive experimental results have demonstrated the superiority of FastBlend. In the future, we intend to integrate FastBlend with other video processing methods to build more powerful video processing tools.

{
  \small
  \bibliographystyle{ieeenat_fullname}
  \bibliography{main}
}



\end{document}